\documentclass{article}

\usepackage[final]{corl_2026} % Uncomment for the camera-ready ``final'' version.
\usepackage{graphicx,subcaption}
\usepackage{amsmath}
\usepackage{enumitem}
\usepackage{adjustbox}
\usepackage[svgnames]{xcolor} % Loads Brown

\usepackage{wrapfig,lipsum,booktabs}
\usepackage{tabularx}
\usepackage{multirow, multicol}
\usepackage{colortbl}
\usepackage{pifont}
\usepackage{makecell}
\usepackage{amssymb}

\usepackage{bm}

\usepackage[most]{tcolorbox}
\usepackage{xcolor}
\usepackage{fontawesome5} % optional; remove icons if you do not want them
\usepackage{changepage}

\definecolor{PromptBlue}{HTML}{1F4E79}
\definecolor{PromptGreen}{HTML}{2E7D32}
\definecolor{PromptPurple}{HTML}{6A1B9A}
\definecolor{PromptGray}{HTML}{555555}
\definecolor{PromptBack}{HTML}{F4F0E5}

\newcommand{\promptsection}[2]{%
  \noindent{\small\bfseries\textcolor{#1}{#2}}\par\vspace{2pt}
}

\newcommand{\promptsubsection}[2]{%
  \hspace*{1.2em}{\footnotesize\bfseries\textcolor{#1}{#2}}\par\vspace{2pt}
}

\newenvironment{promptindent}
  {\begin{adjustwidth}{1.2em}{0pt}\footnotesize}
  {\end{adjustwidth}}

\newcommand{\blfootnote}[1]{%
  \begingroup
  \renewcommand\thefootnote{}%
  \footnote{#1}%
  \addtocounter{footnote}{-1}%
  \endgroup
}
\newcommand{\best}[1]{\textcolor{red}{\bm{#1}}}

\newcommand{\cmark}{\textcolor{green!60!black}{\ding{51}}}
\newcommand{\xmark}{\textcolor{red!70!black}{\ding{55}}}

\title{Monkey See, Can Monkey Do? A Benchmark for Evaluating Robot Skill Learning by Observation}

\author{%
  Weiwei Gu$^{1,}$\thanks{Equal contribution; authors are listed in alphabetical order.}\hspace{3pt}, 
  Anmol Gupta$^{1,*}$,
  Anant Sah$^{1,*}$,
  Ryan Varghese$^1$, \\
  \textbf{
  Lalitha Shreya Vanam$^1$,
  Prabhath Adireddi$^1$,
  Peter Karkus$^2$,
  Nakul Gopalan$^1$}\\
  $^1$Arizona State University, $^2$ Nvidia
}

\begin{document}
\maketitle

%===============================================================================

\begin{abstract}
Learning from Observation (LfO) is a fundamental robotic capability that replicates how humans and animals socially learn from each other.
Beyond its biological parallels, this modality provides a practical solution for data scaling in sample-inefficient and data-starved domains like robotics.
  Recent work has demonstrated promising results in learning manipulation skills from human videos, yet progress in this area remains difficult to assess. Existing methods vary widely in assumptions, hardware choices, and environment setups making it difficult to draw meaningful comparisons and identify advances in the field.
  To address these challenges, we introduce RoboReel: a unified benchmark for evaluating models that learn policies from human videos. 
  RoboReel consists of bundled real-world human demonstration videos, simulated robot trajectories, and evaluation environments on ten manipulation tasks.
  We develop four test suites to evaluate the models' performance on multiple axes, including the robustness to visual distractors and the ability to complete long-horizon tasks.
  We further validate RoboReel by recreating a subset of its tasks in the real world, showing that relative performance trends observed in simulation are preserved in the real world.
  Our benchmark covers learning-from-observation models from different categories, and studies the effectiveness of multiple representation choices in our benchmark evaluation that covers over seven state-of-the-art algorithms (including our VLA based variants) in the field of LfO.
  Finally, we present an analysis of the different types of algorithms showing that long-horizon tasks and tasks with low tolerances are still challenging for current models. Webpage: \href{https://roboreel.github.io}{https://roboreel.github.io}
\end{abstract}

\blfootnote{Correspondence to: \texttt{weiweigu@asu.edu}, \texttt{agupt374@asu.edu}, and \texttt{ng@asu.edu}}

\section{Introduction}
% Opening paragraph to talk about motivation
% Maybe give an example
\begin{figure}
    \centering
    \includegraphics[width=\linewidth]{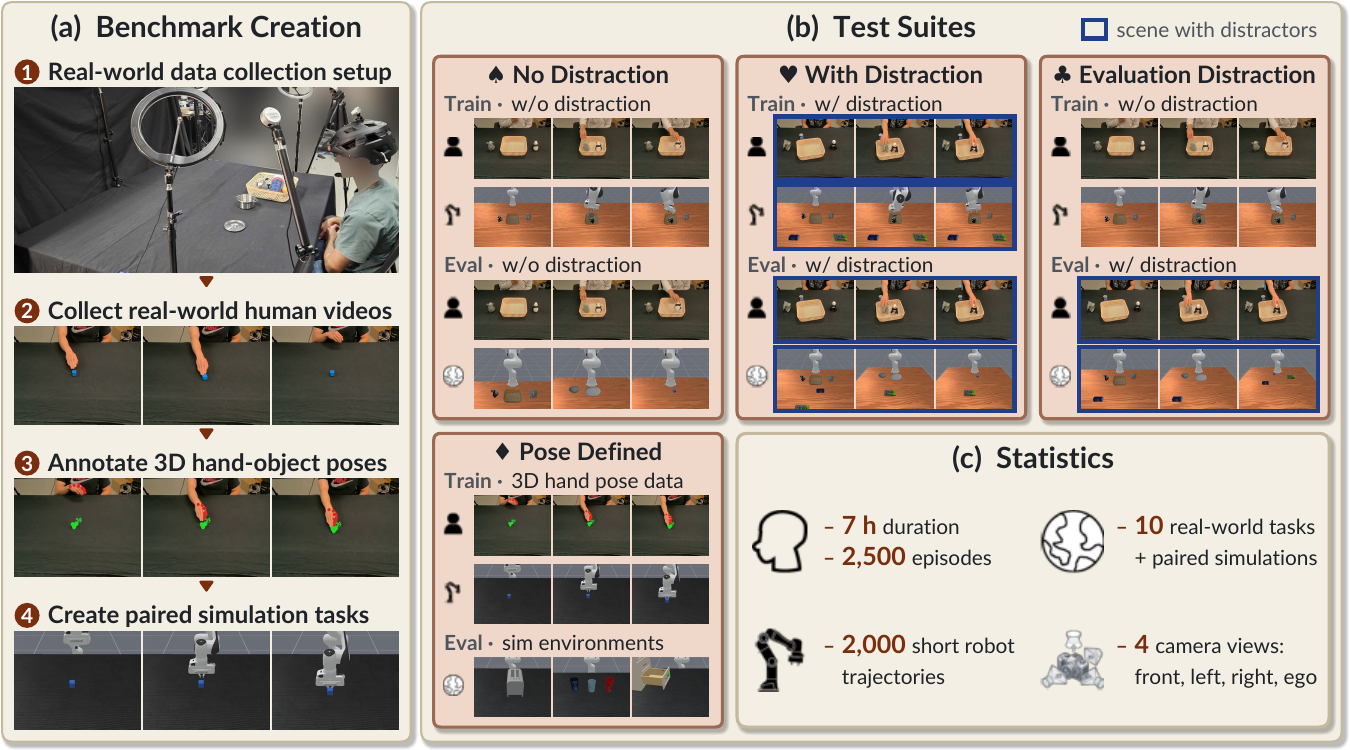}
    \caption{
    % Overview -- RoboReel consists of real world data collected by \ngnote{how many?} people on a calibrated table-top setup for ten tasks. We created a digital-twin simulation domain with paired robot trajectories which has the same objects with matching textures and colors allowing for equitable lateral comparisons for multiple Learning from Observation (LfO) algorithms. The benchmark provides five  test suits -- three of which test for ability to train using human observations with or without distractors and test on settings with or without distractors; one tests for the model's ability to learn one task out of multiple demonstrated; and finally we provide a calibrated test suite for algorithm that use extracted features to demonstrate LfO. The benchmark spans a wide breadth of tasks solved within the LfO community currently with some aspirational goals of long horizon understanding of human videos with distractions.
    \textbf{Overview} -- RoboReel consists of real-world data collected by six participants using a calibrated, tabletop setup across ten distinct tasks with over seven hours of data. To facilitate equitable lateral comparisons of multiple Learning from Observation (LfO) algorithms, we constructed a paired digital-twin simulation environment featuring identical objects with matching textures and colors. The benchmark features four distinct test suites: three evaluate the capability to train on human observations (with or without distractors) and test across varied distraction settings; and the fourth suite offers a calibrated evaluation for feature-extraction-based LfO approaches. RoboReel spans the full breadth of current LfO challenges while introducing aspirational suites for long-horizon understanding of observed tasks in cluttered environments.}
    \label{fig:dataset_description}
\end{figure}

% Motivation for learning from human data

Observational Learning is an important developmental skill for humans~\cite{Bandura1963VICARIOUSRA,Fryling2011UnderstandingOL,Hopper2008ObservationalLI}, other mammals~\cite{Hopper2008ObservationalLI,Mersmann2011SimpleMC} and even birds~\cite{Biederman1986ObservationalLO}.
It is no surprise that this learning paradigm, specifically from human observations has been of interest to the robot learning community with multiple surveys~\cite{lfosurvey, Burnwal2025LearningFO} and state-of-art approaches~\citep{LfOOG, human2robot,bcz,h2r, human2robot,xskill,videoprompt}. 
% Despite this significant work in the area there is a lack of benchmarks that allow us to measure the performance of the state of the art approaches in an apples to apples comparison. 
% % In this work we present a benchmark consisting of paired human and simulated data, offering a reproducible and equitable baseline for assessing different Learning-from-Observation (LfO) algorithms.
% The absence of this reproducibility prevents systematic comparison across methods, as we are not capable of making any lateral (or apples to apples) comparison making it impossible to determine a fair and reproducible state of the art approach making it difficult to establish a reliable state-of-the-art or draw meaningful conclusions about progress in the field
Despite significant progress in this area, the field lacks benchmarks that enable systematic and reproducible evaluation of Learning from Observations (LfO) approaches. The absence of this reproducibility prevents systematic comparison across methods, as we are not capable of making any lateral (or apples to apples) comparison making it difficult to establish a reliable and reproducible state-of-the-art approach or draw meaningful conclusions about progress in the field.

% Learning from observation is a desirable property for robots for two major reasons.
% Firstly, human demonstration videos are a more scalable data source compared to real-world robot trajectories that need calibrated hardware and expert demonstrators~\citep{hommi, rh20t}.
% Collecting real-world robot data requires specialized hardware setup and expert demonstrators to control robots to perform the tasks with teleoperation setups~\citep{hommi, rh20t}. 
% In contrast, human users with little to no experience in robotics can easily provide demonstration videos for different tasks.
% Secondly, compared to on the platform demonstration, teaching by providing observations is a more natural teaching interaction for human users.
% This significantly increases the usability of real-robot applications because human users can teach new knowledge to robots without bringing the robot back to the factory, or setting up the teleoperation infrastructure at their houses to collect robot data.

Given the scaling and biological aspirations, there are significant recent works that focus on learning from human demonstration videos~\citep{bcz,h2r, human2robot,xskill,videoprompt}.
Broadly, these works either convert human videos to language instructions using VLMs~\citep{seedo, physbrain}; or learn some implicit representations from human videos to learn policies~\citep{vid2robot, bcz, rhyme, r3m, mvp, mimicvideo, uniskill}; or utilize accurate 3D human pose annotations in a calibrated space for learning using imitation~\citep{motionTracks, pointpolicy} or in-context learning~\citep{rpx, kat, mimicdroid,videoprompt}.
% These works generally use three different mechanisms to learn from human demonstration videos~\cite{lfosurvey}: \textit{(1) explicitly converting the human videos into language instructions}~\citep{seedo, physbrain}, \textit{(2) learning implicit representations from human videos for policy learning}~\citep{vid2robot, bcz, rhyme, r3m, mvp, mimicvideo, uniskill}, and \textit{(3) directly utilizing 3D pose annotations from human videos as action labels for imitation learning~\citep{motionTracks, pointpolicy} or in-context learning~\citep{rpx, kat, mimicdroid,videoprompt}}.
% Based on task knowledge transfer pathways, these works fall into three broad categories~\cite{lfosurvey}: \textit{(1) task-oriented transfer}~\citep{seedo, physbrain, vid2robot}, \textit{(2) observation-oriented transfer}~\citep{rhyme, r3m, mvp, mimicvideo}, and \textit{(3) action-oriented transfer}~\citep{mimicdroid, pokenet, motionTracks, rpx, pointpolicy, uniskill}.
% With different assumptions on the demonstration videos and the robot environments,
These methods are trained and evaluated with different assumptions on the demonstration videos with varying train and test protocols.
While there are significant efforts in creating datasets and benchmarks for learning from watching~\citep{rh20t, human2robot, mimicdroid},
% RH20T~\cite{rh20t} and Human2Robot~\cite{human2robot} provide paired real-world human videos and robot trajectories, and MimicDroid~\cite{mimicdroid} provides paired simulation robot trajectories and evaluation environments.
% However
 these works do not provide a unified real human dataset with a calibrated and paired test environment with trajectories to train and evaluate LfO models. Crisply, standardized LfO evaluation faces two fundamental challenges: visual fidelity of the scene and the availability of a portable, calibrated environment to test robot performance. 
% with paired real-world human demonstration videos and evaluation environments.

To address this gap, we present RoboReel, a unified dataset and benchmark to train and evaluate LfO models.
% A key challenge in LfO work is the lack of a portable calibrated environment to test the robot performance equitably given a human demonstration. RoboReel solves this challenge by presenting a calibrated real world environment for human demonstrations with a paired simulation environment where a robot's performance can be measured accurately.
RoboReel leverages generative methods to reconstruct accurate 3D meshes of real-world objects, and pairs a calibrated real-world environment for human demonstrations with a matched simulation environment for an equitable evaluation.
The benchmark presents multiple suites for testing a variety of learning algorithms from those that require wrist poses of humans in the observation to those that can account for distractions in the environment. 
The work provides over seven hours of paired and calibrated human data in ten table top tasks from multiple camera angels to measure the capability of LfO algorithms. We demonstrate the benchmark's utility by evaluating several SOTA approaches alongside Vision-Language-Action (VLA) baselines. This evaluation highlights the efficacy of the benchmark while uncovering the performance characteristics and underlying assumptions of different algorithms.
% The simulation environment presented is a digital-twin  preserving visual features of the real-world objects in the simulation environment for testing.
% of a real world human demonstration
% ten tasks encompassing manipulation skills for both rigid and articulated objects.
% For each task, we collect human demonstration videos from both ego-centric and exo-centric camera angles with real-world objects, and create the digital twins of rigid-body objects in the simulation environment, preserving visual features of the real-world objects in the simulation environment.
% Demonstration videos and simulation environments with distraction objects are also provided to analyze the generalizability of these learning-from-observation models.
% Additionally, we provide paired demonstration videos and robot trajectories that comprise sequences of two to three tasks. 
% These demonstrations and trajectories can be used to analyze the models' capability of decomposing and understanding subgoals from the demonstrations.
Our major contributions are as follows:
\begin{enumerate}[leftmargin=5mm,nolistsep]
    \item We introduce a novel learning-from-observation benchmark, providing paired real-world human demonstration videos, robot trajectories, and evaluation environments on \textbf{$10$} tasks. To the best of our knowledge, our work is the first unified benchmark that provides paired training data and evaluation environment using real-world human demonstrations for learning-from-observation.
    \item We use RoboReel to train and evaluate over seven learning-from-observation models with varying representational choices under four different suites, providing a detailed analysis of their effectiveness and the existing gaps in SOTA LfO methods.
    \item  We recreate a subset of RoboReel tasks in the real world and evaluate LfO models in the real world. We observe strong consistency between RoboReel and real-world performance (Pearson $r=0.92$; Spearman $\rho=0.94$), demonstrating that RoboReel can provide a useful indication of the relative performance of LfO models in the real world.
    % \item We train and evaluate over seven learning-from-observation models with varying representational choices. 
    % \item We provide a detailed analysis on the effectiveness of the algorithms and existing gaps in SOTA LfO models.
\end{enumerate}

\section{Related Work}
\label{sec:relatedwork}

\begin{table*}[t]
\centering
\small
\resizebox{\textwidth}{!}{%
\begin{tabular}{c|cccccccc}
\toprule
\multirow{3}{*}{\makecell[c]{\textbf{Dataset}\\\textbf{Name}}} 
& \multirow{3}{*}{\makecell[c]{\textbf{Real-world}\\\textbf{Human}\\\textbf{Vid.}}} 
& \multirow{3}{*}{\makecell[c]{\textbf{Human Demo.}\\\textbf{W. Distractions}}}   
& \multirow{3}{*}{\makecell[c]{\textbf{Multi-view}\\\textbf{Demo.}}} 
& \multirow{3}{*}{\makecell[c]{\textbf{Paired Real-world}\\\textbf{Human Vid.}\\\textbf{Eval Env.}}} 
& \multirow{3}{*}{\makecell[c]{\textbf{Env. w.}\\\textbf{Distractions}}} 
& \multirow{3}{*}{\makecell[c]{\textbf{Multi-view}\\\textbf{Env.}}}
& \multirow{3}{*}{\makecell[c]{\textbf{Articulated} \\\textbf{Objects}}} 
& \multirow{3}{*}{\makecell[c]{\textbf{Long}\\\textbf{Horizon}}} \\
& & & & & & & &\\
& & & & & & & &\\
\midrule
RoboCerebra~\cite{robocerebra}  & \xmark & \xmark & \xmark & \xmark & \cmark & \cmark & \cmark & \cmark\\
LIBERO~\cite{libero}            & \xmark & \xmark & \xmark & \xmark & \cmark & \cmark & \cmark & \cmark\\
RoboCasa~\cite{robocasa}        & \xmark & \xmark & \xmark & \xmark & \cmark & \cmark & \cmark & \cmark\\
VLABench~\cite{vlabench}        & \xmark & \xmark & \xmark & \xmark & \cmark & \cmark & \cmark & \cmark \\
MimicGen~\cite{mimicgen}        & \xmark & \xmark & \xmark & \xmark & \cmark & \cmark & \cmark & \cmark\\
\midrule
EgoVerse~\cite{egoverse} & \cmark & \cmark & \xmark & \xmark  & \xmark & \xmark & \cmark & \cmark \\
Human2Robot~\cite{human2robot}  & \cmark & \cmark & \xmark & \xmark & \xmark & \xmark & \xmark & \cmark \\
RH20T~\cite{rh20t}              & \cmark & \cmark & \cmark & \xmark & \xmark & \xmark & \cmark & \cmark\\
MimicDroid~\cite{mimicdroid}    & \xmark & \cmark & \xmark & \xmark & \cmark & \xmark & \cmark & \xmark\\
\textbf{RoboReel (Ours)} & \cmark & \cmark & \cmark & \cmark & \cmark & \cmark & \cmark & \cmark \\
\bottomrule
\end{tabular}%
}
\caption{\textbf{Benchmark and Dataset Comparison.} RoboReel provides paired real-world human demonstration videos and robot evaluation environments for unified benchmarking of learning from observation methods.  Compared to existing benchmarks, RoboReel enables evaluating LfO methods under different camera views along multiple axes, including robustness and generalization to distractions and performance across diverse tasks.}
\label{tab:benchmark_comparison}
    
\end{table*}

\textbf{Robotics Datasets, Benchmarks, and Simulators.}
The computer vision community has long been interested in understanding human action videos~\citep{ucf101, somethingsomething, activitynet,howto100M,hmdb,assembly101,egoexo4d,epickitchens}. Unfortunately, this data does not allow accurate comparisons for the robots due to the lack of a paired environment to learn or evaluate actions or skills.
We also have significant robotics datasets~\cite{rtx, rh20t, bridge, droid}, benchmarks~\cite{robomme, libero, rlbench, robocerebra, mimicgen}, and simulators~\cite{maniskill, roboverse, robocasa}. These works can contain paired language instructions with robot trajectories and/or a simulated environment.
However, these works lack paired video demonstrations as an input modality, preventing them from evaluating whether an embodied agent can replicate actions seen from human video observations.
RoboMME~\cite{robomme} includes some video-conditioning tasks to evaluate memory of embodied agents but it lacks any real world human data. 
To the best of our knowledge, RH20T~\cite{rh20t} and the MimicDroid benchmark~\cite{mimicdroid} are the closest to our work.
RH20T~\cite{rh20t} contains paired robot trajectories and human demonstration videos while lacking simulated evaluation environments to perform lateral comparisons. 
%to evaluate the performance of the agents.
MimicDroid~\cite{mimicdroid} provides paired robot demonstration videos and evaluation environments, but lacks real world human demonstration inputs.
We address this gap by introducing paired real-world human videos and simulated environments to benchmark robotic learning from observation.

\textbf{Learning Robot Skills from Human Videos.}
Previous LfO surveys have built taxonomies on existing methods by transfer pathways~\cite{lfosurvey} or by algorithmic designs\cite{Burnwal2025LearningFO}.
In this work, we categorize these methods by their representation choice for the input human observation video. 
As illustrated in Fig.~\ref{fig:model_families}, these representation choices fall into three broad categories: \textit{(1) Extracting explicit language representations from videos}~\citep{seedo, physbrain, llmactionplanning}, \textit{(2) Learning implicit latent representation from videos}~\citep{vid2robot, bcz, rhyme, uniskill, pokenet}, and \textit{(3) Learning robot actions from keypoints with in-context-learning or imitation learning}~\citep{motionTracks, rpx, pointpolicy, mimicdroid}.
Explicit methods use VLMs to extract language instructions from human demonstration videos and then use planners or learned motion policies to complete tasks~\citep{seedo, physbrain, llmactionplanning}.
Implicit methods learn to encode the human videos into latent neural skill embeddings, and condition on these learned  latent representations for policy improvement~\citep{vid2robot, uniskill, bcz, rhyme}.
Keypoint-based methods assume a common action space between the robot and the human and extract these actions using keypoints to perform imitation learning~\citep{motionTracks, pointpolicy,kat,rpx}.
To unify the evaluation of existing methods with diverse architectures and assumptions, we introduce a standardized benchmark to compare these model classes on identical tasks.
An extended discussion of related datasets, benchmarks, and LfO methods is provided in Appendix~\ref{appdx:detailed_related_work}.

\section{RoboReel - A Learning from Observation Benchmark}

% Validating learned policies from human observations is challenging forcing the learning from observations

RoboReel provides a standard test bench to evaluate learning from observation (LfO) models. As shown in Table~\ref{tab:benchmark_comparison} while many benchmarks focus on scaling robot data with language conditioning, none include paired human observations for LfO benchmarking.
% While many benchmarks focus on scaling robot data for language-conditioned policies~\citep{robocerebra, libero, vlabench, robocasa, mimicgen}, none include paired human observations. 
The closest comparisons to our benchmark,  MimicDroid~\cite{mimicdroid}, RHT20~\citep{rh20t} and Human2Robot~\citep{human2robot} lack a critical requirement for equitable lateral comparisons of LfO methods -- real world human data paired with simulation environment for systematic policy assessment, which we develop in the following section. Detailed descriptions for the benchmark can be found in Appendix~\ref{appendix:benchmark_details}.

\subsection{Research Categories addressed by the RoboReel}
% \ngnote{Weiwei - Please condense the names of the Training and Evaluation Paradigms here without naming the tasks.}
% Our RoboReel Benchmark features four major test suites with five train-test paradigm pairs.
% As shown in Fig.~\ref{fig:dataset_description}, these test suites include: (1) Learning-from-Observation, (2) Generalization to Distraction, (3) Task Decomposition, and (4) Imitation Learning from Human Videos.
RoboReel features four carefully designed evaluation suites, each targeting a distinct training-evaluation paradigm. The \textbf{No Distraction Suite (ND)} and \textbf{With Distraction Suite (WD)}, evaluate manipulation under clean and distracted conditions, respectively. Distracted setting introduces non-task-related objects randomly sampled from a pre-generated asset pool, while the clean setting contains no such distractors.
The \textbf{Evaluation Distraction Suite (ED)}, examines whether skills learned through observation are robust to observational domain shift introduced at evaluation time. Unlike the distracted setting, distraction objects are absent during training but are introduced in both the human demonstrations and the simulation environment at evaluation time.
Finally we have a \textbf{Pose Defined Suite (PD)}, which examines whether keypoint annotations using a camera calibration setup benefit the LfO paradigm.

\subsection{Task Curation}
\begin{figure}
    \centering
    \includegraphics[width=\linewidth]{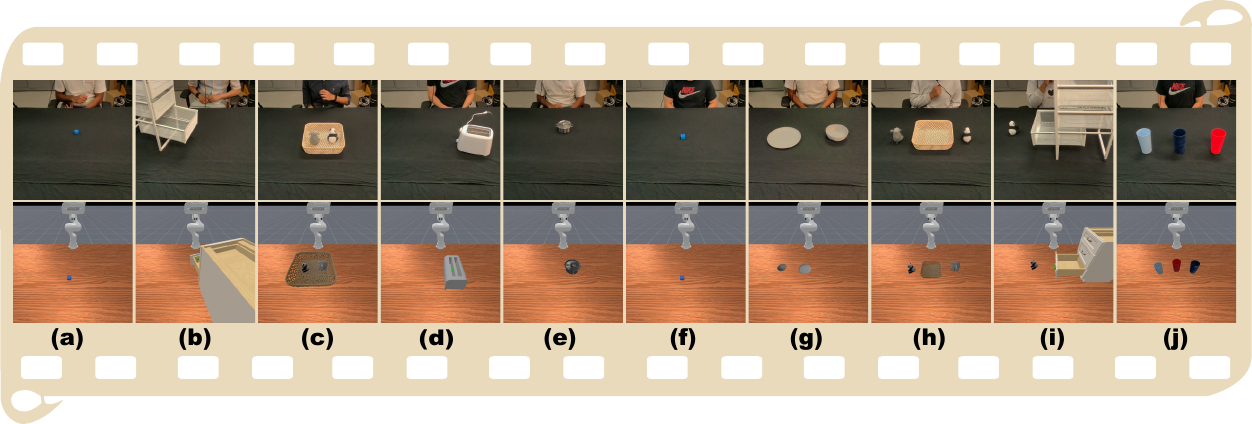}
    \caption{\textbf{Tasks in the RoboReel Benchmark} -- (a) push cube, (b) close drawer, (c) empty basket, (d) press a toaster button, (e) open a lid, (f) pick a cube, (g) place a bowl in a plate, (h) place toys in a basket, (i) place a toy in a drawer and close the drawer, (j) stack three cups. These tasks vary in their horizon, types object interactions, and required tolerances during manipulation.}
    \label{fig:banner}
\end{figure}
To evaluate the learning capability from observations, we design $10$ tasks to test methods across three axes of variation: object type, task horizon, and manipulation precision. These tasks include placing a bowl in a plate, closing a drawer, emptying a basket, opening a lid, picking a cube, pressing a toaster button, pushing a cube, stacking three cups, placing toys in a basket, and placing a toy in a drawer and closing the drawer, as shown in Figure~\ref{fig:banner}. The tasks encompass a wide variety of tabletop manipulation task set that the LfO literature typically evaluates~\cite{pointpolicy, uniskill, seedo}. 
We restrict our scope to goal-based tasks due to two key robotics bottlenecks: inaccurate soft-body simulations and inherent difficulty in validating   non-goal-oriented tasks.

\begin{figure}[t]
    \centering
    \includegraphics[width=\linewidth]{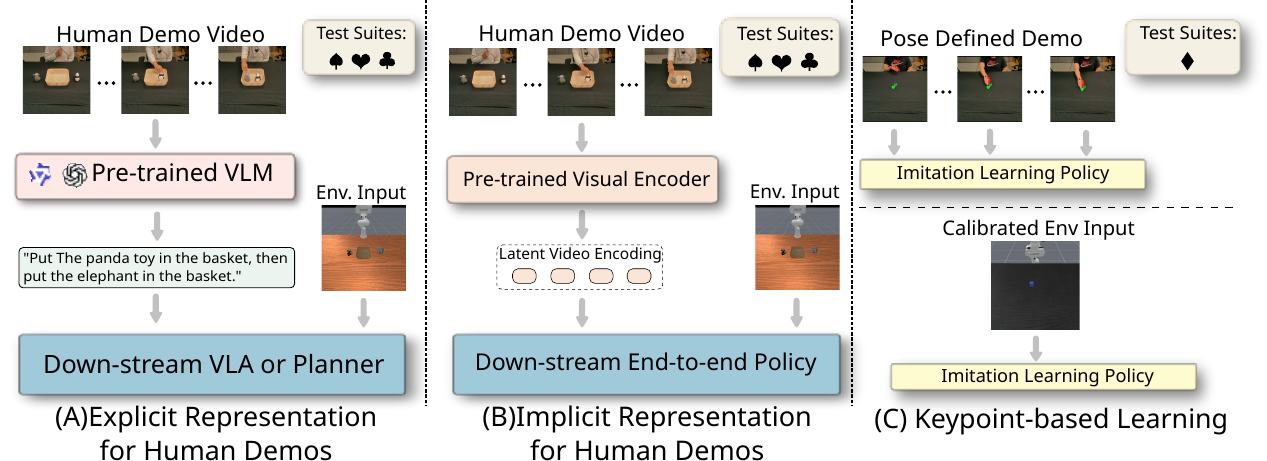}
    \caption{\textbf{Different Model Families for LfO.} We categorize the baseline models we take into consideration by their representational choices to process human demonstration videos. Methods in (A) explicitly generate language instructions with VLMs given videos, and then use a downstream planner or VLA model to generate robot actions. Methods in (B) encode the human demonstrations into a learned latent space which is then used to guide a downstream neural network policy. Models in (C) directly perform imitation learning using human videos and their paired 3D hand poses in training, and directly rollout on an environment that has the same 3D action space in real world.}
    \label{fig:model_families}
\end{figure}

\subsection{Human Demonstration Collection}
To collect human demonstration videos, we construct a tabletop environment in our lab using $18$ rigid-body objects, each with a corresponding digital twin in the simulation environment. The task videos are recorded from four perspectives: three exocentric views capturing the front, left, and right sides of the table using RealSense L515 cameras, and one egocentric view captured by a RealSense D435 camera mounted on a helmet worn by the task performer as shown in Figure~\ref{fig:dataset_description}. Both RGB and depth images are collected from all sensors, with timestamps provided for frame synchronization. 
 Human demonstration videos are collected from six participants with diverse dominant hands, genders, physiques, and demonstration preferences.
We collect $100$ demonstrations per task under clean and distracted conditions respectively, yielding a total of $2{,}000$ human demonstration videos. Table~\ref{tab:benchmark_comparison} provides detailed dataset statistics.

\subsection{Constructing a Corresponding Simulation Environment and Generating Paired data}
% \ngnote{Talk about design choices in the simulation why Maniskill, why the Panda arm, talk about calibration porting. Name the planner. Articulated object models.}
% We build our evaluation environment on ManiSkill~\cite{maniskill}, using a 7-DOF Franka Panda arm.
% To preserve the semantics of the tasks and reduce the visual gap between the real world and simulation, we use SAM3D~\cite{sam3d} to create digital twins for real-world objects. These simulated objects preserve high similarity in visual features, including color, shape, and texture, and have a mesh that closely resembles their physical property in the real world.
% We manually design the scene and success condition for each task to resemble the task settings in the real world.
% Additionally, we also develop a solver for each task to automatically collect robot trajectories for training.

We build our simulated digital twin test bench on ManiSkill~\cite{maniskill} due to its native integration capabilities with PartNet-Mobility dataset~\cite{chang2015shapenet, Xiang_2020_SAPIEN, Mo_2019_CVPR}, GPU-accelerated rendering with ray tracing support and gym-compatible environment interface that enables rapid prototyping of manipulation tasks with varied objects. 
% These properties make ManiSkill particularly well suited for our benchmarking purpose.
We use the 7 DoF Franka Panda arm, chosen for its widespread adoption in  real-world robot learning research. Articulated object assets for drawer and toasters are imported directly from Partnet-mobility dataset~\cite{chang2015shapenet, Xiang_2020_SAPIEN, Mo_2019_CVPR} which provides a rich library of articulated object models of various categories. 

To bridge the visual gap between simulation and the real world, we reconstruct rigid objects as digital twins using SAM3D-objects~\cite{sam3d}. These reconstructed 
meshes preserve high fidelity in color and shapem, and closely approximate the physical geometry of their real-world counterparts. 
We replicate the real-world camera configuration by using ArUco Markers to establish the camera co-ordinate relationships between the three fixed cameras and the table center. This configuration was then applied to the ManuSkill environment allowing camera poses that precisely match their real-world counterparts.
% To replicate the real-world camera configuration in simulation, we place an ArUco 
% marker at the center of the table to establish a fixed reference frame. We compute the transformation from the table center to each of the three cameras, capturing their translation and orientation. These transformations are then directly applied in the ManiSkill environment, where the center of the table of identical dimensions serves as the reference, giving camera poses that precisely match their real-world counterparts. 
This approach ensures visual consistency across domains for benchmarking models that require identical setup during training and testing phases.

To generate training data across tasks, we manually define success conditions that mirror the corresponding real-world configurations. For pick-and-place tasks, we use GraspGen~\cite{murali2025graspgen} to generate plausible candidate grasps, which are then ranked according to constrains including joint movements and end-effector distance. We then use mplib~\cite{guo_mplib} to automatically synthesize 
collision-free robot trajectories across five phases: pre-grasp, grasp, transport, pre-place, and drop. For tasks involving articulated objects such as drawers and 
doors, we define waypoints along the joint motion direction to guide the end-effector in manipulating the target link. Robot skills are labeled successful upon satisfaction of the predefined success goal condition.

% \subsubsection{Benchmark Comparison}
% Table~\ref{tab:benchmark_comparison} provides a detailed comparison between our benchmark and existing benchmarks.
% Most of the previous learning from observation datasets focus on scaling up the number of tasks and human demonstrations. 
% While these provide large scale human video data in real world, due to the difficulty of generalizing to the new environment, without pairing it with a simulated environment and sufficient robot data, it is very difficult to utilize those data to train and evaluate any learning from observation framework.
% Our work focuses on filling this gap by providing paired human demonstration videos, robot trajectories, and an evaluation environment. 
% Additionally, we introduce different levels of distraction in the test environment and the demonstration videos, allowing more profound analysis and understanding on the models' robustness and generalizability on understanding the intention from the demonstration videos and reproducing the actions to achieve the same effect in the evaluation environment.

% \section{Learning from Observation Policies}
% \input{sections/baselines}

\section{Experiments}

\begin{table*}
    \centering
    \scriptsize
    \resizebox{\textwidth}{!}{%
\begin{tabular}{c|c|cccccccccc}
    \toprule
    Model Family
    & Model
    & \makecell{Bowl in\\Plate}
    & \makecell{Close\\Drawer}
    & \makecell{Empty\\Basket}
    & \makecell{Open\\Lid}
    & \makecell{Pick\\Cube}
    & \makecell{Press\\Toaster}
    & \makecell{Push\\Cube}
    & \makecell{Stack\\Cups}
    & \makecell{Toys in\\Basket}
    & \makecell{Toy in\\Drawer} \\
    \midrule

    \multirow{2}{*}{\makecell{\textbf{Explicit}\\\textbf{Representation}}}
    & SeeDo~\cite{seedo}
    & \best{\mathbf{$71.3 \pm 2.2$}} & N/A & \bm{$7.8 \pm 1.9$} & \bm{$54.1 \pm 3.6$} & N/A & N/A & N/A & $0.0 \pm 0.0$ & \bm{$18.7 \pm 1.9$} & N/A \\
    & VLM + $\pi_{0.5}$~\cite{pi05}
    & $1.7 \pm 1.7$
    & \bm{$88.3 \pm 3.3$}
    & $1.7 \pm 1.7$
    & $20.0 \pm 2.9$
    & \bm{$6.7 \pm 1.7$}
    & \bm{$31.7 \pm 3.3$}
    & \bm{$18.3 \pm 1.7$}
    & $0.0 \pm 0.0$
    & $0.0 \pm 0.0$
    & \bm{$16.7 \pm 4.4$} \\
%     & Expert + $\pi_{0.5}$~\cite{pi05}
% & $8.3 \pm 4.4$ & $73.3 \pm 6.7$ & $3.3 \pm 1.7$ & $11.7 \pm 1.7$ & $33.3 \pm 6.0$
% & $1.7 \pm 1.7$ & $0.0 \pm 0.0$ & $1.7 \pm 1.7$ & $0.0 \pm 0.0$ & $0.0 \pm 0.0$ \\

    \midrule

    \multirow{4}{*}{\makecell{\textbf{Implicit}\\\textbf{Representation}}}
    & Vid. Cond. $\pi_{0.5}$~\cite{pi05} (Uniform)
    & $31.7 \pm 3.3$ & $98.3 \pm 1.7$ & $18.3 \pm 6.7$ & \best{$93.3 \pm 4.4$} & \best{$55.0 \pm 5.8$}
    & $83.3 \pm 3.3$ & $78.3 \pm 6.0$ & \best{$11.7 \pm 3.3$} & $20.0 \pm 2.9$ & $16.7 \pm 7.3$ \\
    & Vid. Cond. $\pi_{0.5}$~\cite{pi05} (Keyframes)
        & \bm{$46.7 \pm 6.0$}
        & \best{$100.0 \pm 0.0$}
        & \best{$20.0 \pm 2.9$}
        & $73.3 \pm 1.7$
        & $23.3 \pm 7.3$
        & \best{$88.3 \pm 6.7$}
        & \best{$100.0 \pm 0.0$}
        & $8.3 \pm 3.3$
        & \best{$21.7 \pm 3.3$}
        & \best{$20.0 \pm 7.6$}\\
    % & RHyME~\cite{rhyme}
    % & & & & & & & & & & \\
    & Vid2Robot~\cite{vid2robot}
    & $1.7 \pm 1.7$
    & $21.7 \pm 3.3$
    & $3.3 \pm 1.7$
    & $3.3 \pm 3.3$
    & $1.7 \pm 1.7$
    & $21.7 \pm 6.0$
    & $5.0 \pm 2.9$
    & $3.3 \pm 1.7$
    & $0.0 \pm 0.0$
    & $0.0 \pm 0.0$\\
    & UniSkill~\cite{uniskill}
    & $0.0 \pm 0.0$
    & $66.7 \pm 8.8$
    & $0.0 \pm 0.0$
    & $8.3 \pm 1.7$
    & $10.0 \pm 2.9$
    & $25.0 \pm 10.0$
    & $0.0 \pm 0.0$
    & $0.0 \pm 0.0$
    & $0.0 \pm 0.0$
    & $0.0 \pm 0.0$\\

    % \midrule\midrule

    % \multirow{1}{*}{\makecell{\textbf{Imitation Learning}}}
    % & Point-Policy
    % & & & & & & & & & & \\

    \bottomrule
\end{tabular}%
    }
    \caption{\textbf{Experiment results on the ND test suite ($\spadesuit$).} 
    % Methods using explicit representations convert human demonstration videos into language instructions, while implicit methods learn latent representations from the videos to directly predict robot actions. 
    Models are trained and evaluated without distractions. This configuration studies whether models can perform LfO under the vanilla setting of no distractions. The two video-conditioned $\pi_{0.5}$ policies, which are implicit, i.e., do not use intermediate language descriptions, achieve similar high performance.}
    \label{tab:no_distraction_training_no_distraction_eval_per_task}
\end{table*}

\begin{table*}
    \centering
    \scriptsize
    \resizebox{\textwidth}{!}{%
        \begin{tabular}{c|c|cccccccccc}
        \toprule
        Model Family
        & Model
        & \makecell{Bowl in\\Plate}
        & \makecell{Close\\Drawer}
        & \makecell{Empty\\Basket}
        & \makecell{Open\\Lid}
        & \makecell{Pick\\Cube}
        & \makecell{Press\\Toaster}
        & \makecell{Push\\Cube}
        & \makecell{Stack\\Cups}
        & \makecell{Toys in\\Basket}
        & \makecell{Toy in\\Drawer} \\
        \midrule
    
        \multirow{1}{*}{\makecell{\textbf{Explicit}\\\textbf{Representation}}}
        & SeeDo~\cite{seedo}
        & \bm{$66.3 \pm 0.4 $} & N/A & $4.9 \pm 2.0$ & \bm{$37.8 \pm 3.1$} & N/A & N/A & N/A & $0.0 \pm 0.0$ & \best{$11.1 \pm 2.3$} & N/A \\
        & VLM + $\pi_{0.5}$~\cite{pi05}
        & $30.0 \pm 7.6$
        & $63.3 \pm 4.4$
        & \bm{$13.3 \pm 1.7$}
        & $18.3 \pm 4.4$
        & \bm{$5.0 \pm 5.0$}
        & $13.3 \pm 1.7$
        & $15.0 \pm 5.0$
        & $0.0 \pm 0.0$
        & $1.7 \pm 1.7$
        & $5.0 \pm 2.9$ \\
    
        \midrule
    
        \multirow{4}{*}{\makecell{\textbf{Implicit}\\\textbf{Representation}}}
        & Vid. Cond. $\pi_{0.5}$~\cite{pi05} (Uniform)
        & $86.7 \pm 4.4$
        & \best{$86.7 \pm 3.3$}
        & \best{$25.0 \pm 2.9$}
        & $73.3 \pm 10.1$
        & $13.3 \pm 6.0$
        & $45.0 \pm 7.6$
        & \best{$71.7 \pm 8.3$}
        & $5.0 \pm 0.0$
        & \bm{$5.0 \pm 2.9$}
        & \best{$23.3 \pm 3.3$} \\
        & Vid. Cond. $\pi_{0.5}$~\cite{pi05} (Keyframes)
        & \best{$88.3 \pm 7.3$}
        & $85.0 \pm 5.0$
        & $21.7 \pm 7.3$
        & \best{$85.0 \pm 2.9$}
        & \best{$48.3 \pm 6.0$}
        & \best{$56.7 \pm 6.0$}
        & $21.7 \pm 6.0$
        & \best{$8.3 \pm 4.4$}
        & $3.3 \pm 1.7$
        & $18.3 \pm 1.7$\\
        % & RHyME~\cite{rhyme}
        % & & & & & & & & & & \\
        & Vid2Robot~\cite{vid2robot}
        & $6.7 \pm 4.4$
        & $10.0 \pm 5.8$
        & $3.3 \pm 3.3$
        & $0.0 \pm 0.0$
        & $0.0 \pm 0.0$
        & $30.0 \pm 5.0$
        & $8.3 \pm 3.3$
        & $0.0 \pm 0.0$
        & $0.0 \pm 0.0$
        & $0.0 \pm 0.0$ \\
        & UniSkill~\cite{uniskill}
        & $0.0 \pm 0.0$
        & $56.7 \pm 13.6$
        & $10.0 \pm 2.9$
        & $21.7 \pm 6.7$
        & $1.7 \pm 1.7$
        & $21.7 \pm 7.3$
        & $8.3 \pm 1.7$
        & $0.0 \pm 0.0$
        & $0.0 \pm 0.0$
        & $0.0 \pm 0.0$\\
        \bottomrule
    \end{tabular}%
    }   
    \caption{\textbf{Experiment results on the WD test suite ($\heartsuit$).} Models are trained and evaluated with distractions. This configuration studies whether models can perform LfO in the presence of distractions. We again observe video-conditioned $\pi_{0.5}$ models substantially outperform the other methods, with comparable performance between uniform and keyframe sampling.}
    \label{tab:with_distraction_training_with_distraction_eval_per_task}
\end{table*}

% \begin{figure}
%     \centering
%     \resizebox{0.4\textwidth}{!}{%
%     \includegraphics[width=\linewidth]{figures/lfo_fig3.jpg}
%     }
%     \caption{This figure demonstrates the results of the decomposition task suite. We report results for $3$ target tasks.}
%     \label{fig:decomp_results_and_point_policy}
% \end{figure}

In this section, we systematically evaluate different learning-from-observation methods using RoboReel. 
To show that the performance on RoboReel indicates real-world LfO performance, we also conduct real-world experiments on three representative tasks.
We first describe our baselines in Sec.~\ref{sec:baselines}, followed by the experimental results and analysis in Secs.~\ref{sec:results} and~\ref{sec:analysis}, respectively.
Detailed descriptions of the experiment protocols are provided in Appendix~\ref{appendix:benchmark_details}, and additional experiment results and analysis are provided in Appendix~\ref{appdx:detailed_results}.

\subsection{Baselines}
\label{sec:baselines}

We have categorized our baselines into three families as described in Sec.~\ref{sec:relatedwork} and shown in Figure~\ref{fig:model_families}.

\textbf{Explicit Language Representation from Human Demos.}
For this category, we evaluate SeeDo~\cite{seedo} by extracting the high-level plan generated by a pre-trained VLM planner. As SeeDo requires no training, we report its With Distraction results are equivalent to its Evaluation Distraction results. The action vocabulary here is also restricted to pick-and-place, so only the five pick-and-place tasks that fall within its capabilities are reported. We additionally implement a VLA-based explicit model using Qwen3-VL-4B \cite{qwen3} to extract language commands from demonstration videos, which are used to fine-tune $\pi_{0.5}$~\cite{pi05}. To assess the quality of VLM-generated language guidance, we also train a $\pi_{0.5}$ baseline using expert language annotations.

\textbf{Learning Implicit Representation from Human Demos.}
For this line of work, we train and test established baselines UniSkill~\cite{uniskill} and Vid2Robot~\cite{vid2robot}. We additionally augment $\pi_{0.5}$~\cite{pi05} to accept human video inputs by encoding demonstration frames into visual tokens, using two selection strategies: uniform frame sampling and a BPP-inspired keyframe selection method~\cite{bpp} that weights frames by proximity to the nearest keyframe.

\textbf{Pose Defined LfO.}
We evaluate PointPolicy~\cite{pointpolicy} as our baseline for this category. We report results for five tasks here, with complete results provided in Appendix~\ref{appdx:detailed_results}.
Architecture details and training details are provided in Appendix~\ref{appendix:arch:vapi05} and Appendix~\ref{appendix:training_details}, respectively.

\begin{table*}
    \centering
    \scriptsize
    \resizebox{\textwidth}{!}{%
    \begin{tabular}{c|c|cccccccccc}
    \toprule
    Model Family
    & Model
    & \makecell{Bowl in\\Plate}
    & \makecell{Close\\Drawer}
    & \makecell{Empty\\Basket}
    & \makecell{Open\\Lid}
    & \makecell{Pick\\Cube}
    & \makecell{Press\\Toaster}
    & \makecell{Push\\Cube}
    & \makecell{Stack\\Cups}
    & \makecell{Toys in\\Basket}
    & \makecell{Toy in\\Drawer} \\
    \midrule

    \multirow{2}{*}{\makecell{\textbf{Explicit}\\\textbf{Representation}}}
    & SeeDo~\cite{seedo}
        & \best{$66.3 \pm 0.4 $} & N/A & $4.9 \pm 2.0$ & \bm{$37.8 \pm 3.1$} & N/A & N/A & N/A & $0.0 \pm 0.0$ & \bm{$11.1 \pm 2.3$} & N/A \\
    & VLM + $\pi_{0.5}$~\cite{pi05}
    & $6.7 \pm 3.3$
    & $61.7 \pm 3.3$
    & $5.0 \pm 5.0$
    & $6.7 \pm 1.7$
    & $11.7 \pm 4.4$
    & $15.0 \pm 5.0$
    & $6.7 \pm 1.7$
    & $0.0 \pm 0.0$
    & $0.0 \pm 0.0$
    & $0.0 \pm 0.0$ \\
%     & Expert + $\pi_{0.5}$~\cite{pi05}
% & $5.0 \pm 2.9$ & $43.3 \pm 11.7$ & $5.0 \pm 5.0$ & $25.0 \pm 2.9$ & $28.3 \pm 8.8$
% & $0.0 \pm 0.0$ & $0.0 \pm 0.0$ & $0.0 \pm 0.0$ & $0.0 \pm 0.0$ & $0.0 \pm 0.0$ \\

    \midrule

    \multirow{4}{*}{\makecell{\textbf{Implicit}\\\textbf{Representation}}}
    & Vid. Cond. $\pi_{0.5}$~\cite{pi05} (Uniform)
    & $31.7 \pm 4.4$ & $58.3 \pm 8.3$ & $18.3 \pm 4.4$ & $61.7 \pm 4.4$ & $25.0 \pm 5.8$
    & $15.0 \pm 5.0$ & $6.7 \pm 4.4$ & $5.0 \pm 2.9$ & $11.7 \pm 4.4$ & $6.7 \pm 1.7$ \\
    & Vid. Cond. $\pi_{0.5}$~\cite{pi05} (Keyframe)
        & $25.0 \pm 5.0$
        & \best{$70.0 \pm 2.9$}
        & \best{$21.7 \pm 4.4$}
        & \best{$63.3 \pm 4.4$}
        & \best{$30.0 \pm 5.0$}
        & $26.7 \pm 4.4$
        & \best{$31.7 \pm 3.3$}
        & \best{$11.7 \pm 4.4$}
        & \best{$18.3 \pm 3.3$}
        & $6.7 \pm 1.7$ \\
    % & RHyME~\cite{rhyme}
    % & & & & & & & & & & \\
    & Vid2Robot~\cite{vid2robot}
    & $0.0 \pm 0.0$
    & $23.3 \pm 1.7$
    & $5.0 \pm 2.9$
    & $0.0 \pm 0.0$
    & $0.0 \pm 0.0$
    & $23.3 \pm 9.3$
    & $6.7 \pm 3.3$
    & $0.0 \pm 0.0$
    & $0.0 \pm 0.0$
    & $0.0 \pm 0.0$ \\
    & UniSkill~\cite{uniskill}
    & $0.0 \pm 0.0$
    & $63.3 \pm 14.8$
    & $0.0 \pm 0.0$
    & $18.3 \pm 1.7$
    & $1.7 \pm 1.7$
    & \best{$28.3 \pm 8.8$}
    & $0.0 \pm 0.0$
    & $0.0 \pm 0.0$
    & $0.0 \pm 0.0$
    & $0.0 \pm 0.0$ \\

    % \midrule\midrule

    % \multirow{1}{*}{\makecell{\textbf{Imitation Learning}}}
    % & Point-Policy
    % & & & & & & & & & & \\

    \bottomrule
\end{tabular}%
    }
    \caption{\textbf{Experiment results on the ED test suite ($\clubsuit$).} This evaluation's demonstrations and environments contain distractions, while the training data does not. This configuration evaluates whether models can robustly perform LfO and generalize to distracted settings unseen during training. Video-augmented $\pi_{0.5}$ with keyframes achieves the best performance on most tasks.}
    \label{tab:no_distraction_training_with_distraction_eval_per_task}
    
\end{table*}

% \begin{figure}
%     \centering
%     \includegraphics[width=0.33\linewidth]{figures/keypoint_bar.pdf}
%     \caption{This figure presents the experiment results on the keypoint-based test suite($\bigstar$).}
%     \label{fig:keypoint}
% \end{figure}

% \begin{figure}
    % \begin{minipage}{0.33\linewidth}

\subsection{Results}
\label{sec:results}

\textbf{No Distraction Test Suite.} 
Table~\ref{tab:no_distraction_training_no_distraction_eval_per_task} shows the performance on the \textbf{ND} Suite. 
On this test suite, video-conditioned policies with implicit task representations perform strongest overall.
Our two baseline variants of video-conditioning $\pi_{0.5}$ models outperform other baselines significantly across most tasks.
The performance gap is especially obvious in the long-horizon tasks that involve multiple sub-tasks, including Empty Basket, Stack Cups, Toys in Basket, and Toy in Drawer.
% \ngnote{analysis -  This result demonstrates that the internet-scale pre-training of the VLAs helps them quickly understand the correspondence between the human video features and the robot actions. }
Despite having access to action primitives that can access the ground truth locations of the objects, SeeDo~\cite{seedo} does not perform well on multi-step pick-and-place tasks such as Empty Basket and Stack Cups.
% \ngnote{analysis - This shows VLM's limited ability to generate proper executable plans from the human demonstration videos. this compared to picking keyframes}
\begin{wrapfigure}{r}{0.45\textwidth}
    \centering
    \includegraphics[width=\linewidth]{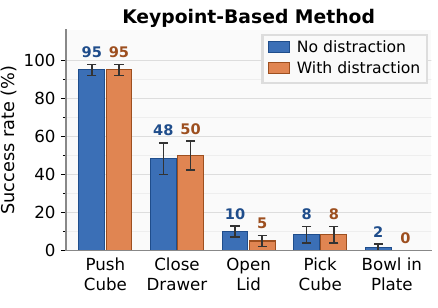}
    \caption{\textbf{Experiment results of PointPolicy which only trains on human data on five tasks of the pose-defined test suite ($\diamondsuit$).} PointPolicy performs well on Push Cube and Close Drawer and remains robust to distractions, but struggles on the three tasks requiring precise grasping, suggesting that keypoints on human data can capture the overall trajectory behavior but provide insufficient grasp specifications.}
    \label{fig:keypoint}
\end{wrapfigure}

\textbf{With Distraction Test Suite. }
Table~\ref{tab:with_distraction_training_with_distraction_eval_per_task} describes the models' performance on the \textbf{WD} Suite.
Similar to the \textbf{ND} suite, the two video-conditioned $\pi_{0.5}$ variants achieve the overall best performance.
Although the \textbf{WD} suite evaluates models' performance with in-distribution settings, models from both families suffer a performance drop on the majority of the tasks.

\textbf{Evaluation Distraction Test Suite. }
Table~\ref{tab:no_distraction_training_with_distraction_eval_per_task} describes the performance on the \textbf{ED} Suite.
In this test suite, we find video-conditioned $\pi_{0.5}$ with keyframes achieves the overall best performance on most tasks.
We also find a similar trend that implicit methods outperforming explicit methods on all tasks with significance.
Because this \textbf{ED} suite is by design more challenging than the previous two task suites, most models cannot achieve a similar performance to the other task suites on most tasks.

\textbf{Pose Defined Test Suite.}
In this test suite, we report the results of PointPolicy~\cite{pointpolicy} on the five selected tasks in Fig.~\ref{fig:keypoint} with the rest of the results in the Appendix~\ref{appdx:detailed_results}.
We find that PointPolicy is able to capture the behaviors of the human demonstrators from the keypoint data, achieving a high performance in the Push Cube and the Close Drawer task. 
Additionally, this method displays significant robustness towards distractions, achieving a similar performance under the environment with and without distraction. 
However, this method does not perform on the other three tasks that involve grasping, regardless of the distraction setting.

\begin{wrapfigure}{r}{0.45\textwidth}
    \centering
    \includegraphics[width=\linewidth]{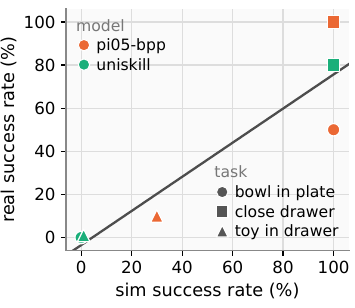}
    \caption{\textbf{Sim. vs. Real Results on Selected Tasks.} Success rates in simulation and the real world for video-conditioned $\pi_{0.5}$ and UniSkill across three representative tasks. High Pearson ($r=0.92$) and Spearman ($\rho=0.94$) correlations show that relative performance trends are largely preserved in the real world robot results.}
    \label{fig:sim_vs_real}
\end{wrapfigure}

\textbf{Real-World Evaluation.}
We evaluate video-conditioned $\pi_{0.5}$ and UniSkill on the three selected tasks.
Fig.~\ref{fig:sim_vs_real} presents the simulation and real-world performance across the six model-task pairs.
Simulation and real-world performance exhibit consistent relative trends, with a Pearson correlation of $r=0.92$ and a Spearman rank correlation of $\rho=0.94$.
Overall, the results demonstrate strong rank consistency between simulation and real-world performance.

% \begin{figure*}[t]
%     \centering

%     \begin{minipage}[t]{0.40\textwidth}
%         \centering
%         \resizebox{!}{4.8cm}{
%             \includegraphics{figures/keypoint_bar.pdf}
%         }
%         \caption{This figure presents the experiment results of PointPolicy on the keypoint-based test suite ($\diamondsuit$) on five tasks. A different individual model is trained for each task.}
%         \label{fig:keypoint}
%     \end{minipage}
%     \hfill
%     \begin{minipage}[t]{0.54\textwidth}
%         \centering
%         \resizebox{!}{5.2cm}{
%             \includegraphics{figures/sim_vs_real.png}
%         }
%         \caption{\change{Success rates in simulation and real-robot for video-conditioned $\pi_{0.5}$ and UniSkill across three representative tasks. Colors indicate models and marker shapes indicate tasks. The line shows the linear fit between simulation and real-world success rates.}}
%         \label{fig:sim_vs_real}
%     \end{minipage}

% \end{figure*}

\subsection{Analysis}
\label{sec:analysis}

We analyze our results by answering the following key research questions:

\textbf{Q1. Which model family performs the best?} \textbf{Across all ten tasks and every test suite, video-conditioned $\pi_{0.5}$ variants achieve the strongest overall performance}, which we attribute to the large-scale pretraining of VLA models that enables better correspondence between human video features and robot actions. While the difference between uniform-sampled and keyframe-based $\pi_{0.5}$
$\pi_{0.5}$ is marginal in the ND and WD suites, where training and evaluation share the same distraction conditions, the gap becomes stark in the ED suite, where \textbf{keyframe-based $\pi_{0.5}$} outperforms uniform sampling in $7$ out of $10$ tasks. We attribute this to keyframes capturing task-relevant moments with significance in the demonstration, making the policy more robust to the visual domain shift introduced at evaluation time in the ED suite. In contrast, uniform sampling includes many redundant or uninformative frames that may act as noise when the evaluation distribution shifts.

\textbf{Q2. Does the task length affect the performance?} \textbf{Task horizon has a clear negative affect on success rate across all model families.} Long-horizon tasks yield lowest success rate across all suites. These tasks require the robot to execute multiple sequential sub-goals correctly, where failure at any intermediate step results to overvall task failure. In contrast, short-horizon tasks such as Push Cube, Pick Cube, and Close Drawer see substantially higher success rates across all model families. Even the strongest baseline, video conditioned $\pi_{0.5}$ variants struggles on long horizon tasks with success remaining below 25\%. 

\textbf{Q3. Do VLAs benefit more from explicit natural language guidance?}
We hypothesized that the internet-scale pretraining enables VLMs to understand simple task demonstration videos, and can correctly summarize the natural language instructions regardless of the distraction objects.
Additionally, we expected VLAs to perform better with the language modality as these VLAs are extensively pre-trained with instruction following.
However, \textbf{from our experiment results across three test suites, we find that implicit representation consistently outperforms explicit representation.}
Although both VLMs and VLAs go through internet-scale pre-training with image and text data, we hypothesize the extra exposure to the robot trajectory data helps VLAs to overcome VLMs' weakness in visual-temporal reasoning~\cite{temporalVQA}.

\textbf{Q4. Can models learn tasks requiring lower tolerances manipulation?} We identify Press Toaster and Stack Cups as two tasks have low tolerances between objects during manipulation. For Press Toaster, models achieve relatively good performance in the ND suite with $\pi_{0.5}$ variant performing over $88\%$,  but performance degrades consistently from the ND suite to the WD suite and further to the ED suite across all model families. We attribute this degradation to the small target area of the toaster button, where distractor objects in the scene introduce visual clutter that confuses the vision encoder, making it harder to localize the precise interaction point. For Stack Cups, performance is uniformly low across all suites and all models, as this task not only requires precise placement but is also long-horizon, compounding the difficulty.

\textbf{Q5. Do calibrated keypoints help in learning from observations?} From our experiments with PointPolicy~\cite{pointpolicy}, we find that while keypoint annotations help the model learn the overall trajectory behavior, this does not translate to consistent performance gains across tasks. As shown in Figure 4, PointPolicy achieves strong results on tasks such as Push Cube and Close Drawer, but performance drops sharply on tasks requiring precise grasping such as Pick Cube, Open Lid, and Bowl in Plate. We attribute this to the inherent difficulty of deriving grasp configurations purely from keypoints, especially on objects like bowls and handles. Precise grasp specifications without images is causing many grasp failures in our observations.

\textbf{Q6. Is performance in RoboReel predictive of real-world LfO performance?}
Our real-world evaluation shows high Spearman and Pearson correlations, indicating that both model rankings and relative performance trends are largely preserved between simulation and the real world.
A gap in absolute success rates still exists because RoboReel does not fully replicate the physical properties of real-world objects and includes only a limited set of tasks. Future work can improve simulation fidelity and expand the diversity of tasks to further reduce this gap.

\textbf{Q7. What are the fundamental limitations of current LfO methods?} Our analysis across five research questions reveals consistent gaps in the current LfO paradigm. First, all model families struggle with long-horizon tasks, where compounding errors across sequential sub-goals remain a critical bottleneck. Second, precise manipulation tasks are hard to learn, especially in cluttered environments.
Finally, while large-scale VLA pretraining provides a strong foundation, long-horizon tasks and tasks requiring precise manipulation remain unsolved across all evaluated methods.

\section{Limitations}
A significant limitation of our approach is the use of goal-based tasks. Humans can perform non-Markovian and dynamic skills whose goals cannot be tested at a single time point, but this is a wider challenge within robotics.
Our benchmark also evaluates a limited set of tasks, and future work should expand the task diversity to better capture the range of skills encountered in real-world LfO.
We also do not have any soft objects because their physics are harder to simulate.  
Another major limitation of our approach is the lack of bimanual tasks. Given most approaches in LfO work on single arm domains we restricted ourselves to tasks that use a single arm. We note that the presence of multiple arms would make action generation significantly challenging for existing models. 
The reconstructed objects do not replicate their exact physical properties, such as material, mass, and friction. 
We were also unable to reconstruct the articulated objects, and instead used the closest matching models from an existing asset library. These limitations are partly inherited from current simulation and object-reconstruction tools.
Finally, future work is needed to investigate whether learning from human demonstrations reduces the need for robot trajectories.

\section{Conclusion}
In conclusion, we contribute RoboReel, a benchmark to evaluate existing Learning-from-Observation frameworks with real-world human demonstrations.
We manually curate $10$ tasks, collect $2,000$ real-world human demonstrations from multiple angles, and develop paired evaluation environments in the simulation.
Using the $4$ task suites from RoboReel, we study over seven SOTA models from various model families, and analyze  their performance on learning from observation under different settings.
Through our real-world experiments, we demonstrate that the relative performance trends observed in RoboReel are largely preserved in the real world, supporting its use as an indicator of real-world LfO model performance.
To the best of our knowledge, RoboReel is the first benchmark that introduces a shared evaluation protocol for these different model families.
We plan to continuously develop RoboReel to increase the scale and variety of the data over time. 
% We also plan to include dynamic tasks and non-goal-based tasks, and expand the target embodiments of learning in the future.

\newpage

\section*{Acknowledgement}
This material is based upon work supported by the Air Force Office of Scientific Research under award number FA9550-24-1-0239. Any opinions, findings, and conclusions or recommendations expressed in this material are those of the author(s) and do not necessarily reflect the views of the United States Air Force. We also want to thank our anonymous reviewers for providing great feedback and making the work stronger.

%===============================================================================

% \clearpage
% The acknowledgments are automatically included only in the final and preprint versions of the paper.
% \acknowledgments{If a paper is accepted, the final camera-ready version will (and probably should) include acknowledgments. All acknowledgments go at the end of the paper, including thanks to reviewers who gave useful comments, to colleagues who contributed to the ideas, and to funding agencies and corporate sponsors that provided financial support.}

%===============================================================================

% no \bibliographystyle is required, since the corl style is automatically used.
\bibliography{example}  % .bib
\clearpage
\appendix

\section*{Contributions}
\textbf{Weiwei Gu.} Implemented the models, evaluated baselines, organized the real-world data collection, wrote the appendix, contributed to writing the paper and supported the real robot experiments.

\textbf{Anmol Gupta.} Primarily developed the benchmark, designed the simulated tasks, developed the simulation data-collection pipeline, supported real-world data collection, and served as the primary writer of the paper.

\textbf{Anant Sah.} Conducted the real-world robot experiments, contributed to real-world data collection, and supported the implementation of baseline methods.

\textbf{Ryan Varghese.} Collected the real-world human demonstration data.

\textbf{Lalitha Shreya Vanam.} Collected the real-world human demonstration data.

\textbf{Prabhath Adireddi.} Collected the real-world human demonstration data and assisted with collecting the real-world robot data.

\textbf{Peter Karkus.} Provided guidance on the problem formulation and experimental design.

\textbf{Nakul Gopalan.} Supervised the project and contributed to writing the paper.

\section{Additional Experiment Results}
\label{appdx:detailed_results}

\begin{wraptable}[12]{r}{0.48\textwidth}
\centering
\scriptsize
\begin{tabular}{llcc}
\toprule
\textbf{Model} & \textbf{Task} & \textbf{Planner Data} & \textbf{Teleop Data} \\
\midrule
\multirow{3}{*}{Vid. $\pi_{0.5}$-BPP}
    & Bowl   & $\mathbf{1.00 \pm 0.00}$ & $0.87 \pm 0.09$ \\
    & Drawer & $1.00 \pm 0.00$ & $0.97 \pm 0.03$ \\
    & Toy    & $0.53 \pm 0.09$ & $\mathbf{0.73 \pm 0.03}$ \\
\midrule
\multirow{3}{*}{UniSkill}
    & Bowl   & $0.03 \pm 0.03$ & $0.10 \pm 0.06$ \\
    & Drawer & $1.00 \pm 0.00$ & $1.00 \pm 0.00$ \\
    & Toy    & $0.00 \pm 0.00$ & $0.00 \pm 0.00$ \\
\bottomrule
\end{tabular}
\caption{Success rates of policies trained using planner-generated versus teleoperated robot trajectories. Bold indicates the higher-performing training source for each model-task pair.}
\label{tab:planner_vs_teleop}
\end{wraptable}

\subsection{Additional Results for PointPolicy}
Across the five extended tasks (empty basket, stack cups, toaster, toy drawer close, and toys in basket), the trained Point Policy achieves 0\% success under the simulator's native success criterion. The policy nonetheless exhibits directionally correct behavior on several tasks, approaching the toys in toys in basket and reaching the drawer in toy drawer close. Two factors contribute to this performance gap.

First, the extended tasks have substantially longer horizons and require multiple sequential sub-skills: approach, grasp, lift, transport, and release. By contrast, the previously evaluated tasks were largely single-stage motions such as a planar push or a single grasp. The policy reliably reproduces the initial approach phase observed in the demonstrations but frequently fails to execute the subsequent manipulation stages needed to satisfy the environment's success condition.

\begin{figure}
    \centering
    \includegraphics[width=\linewidth]{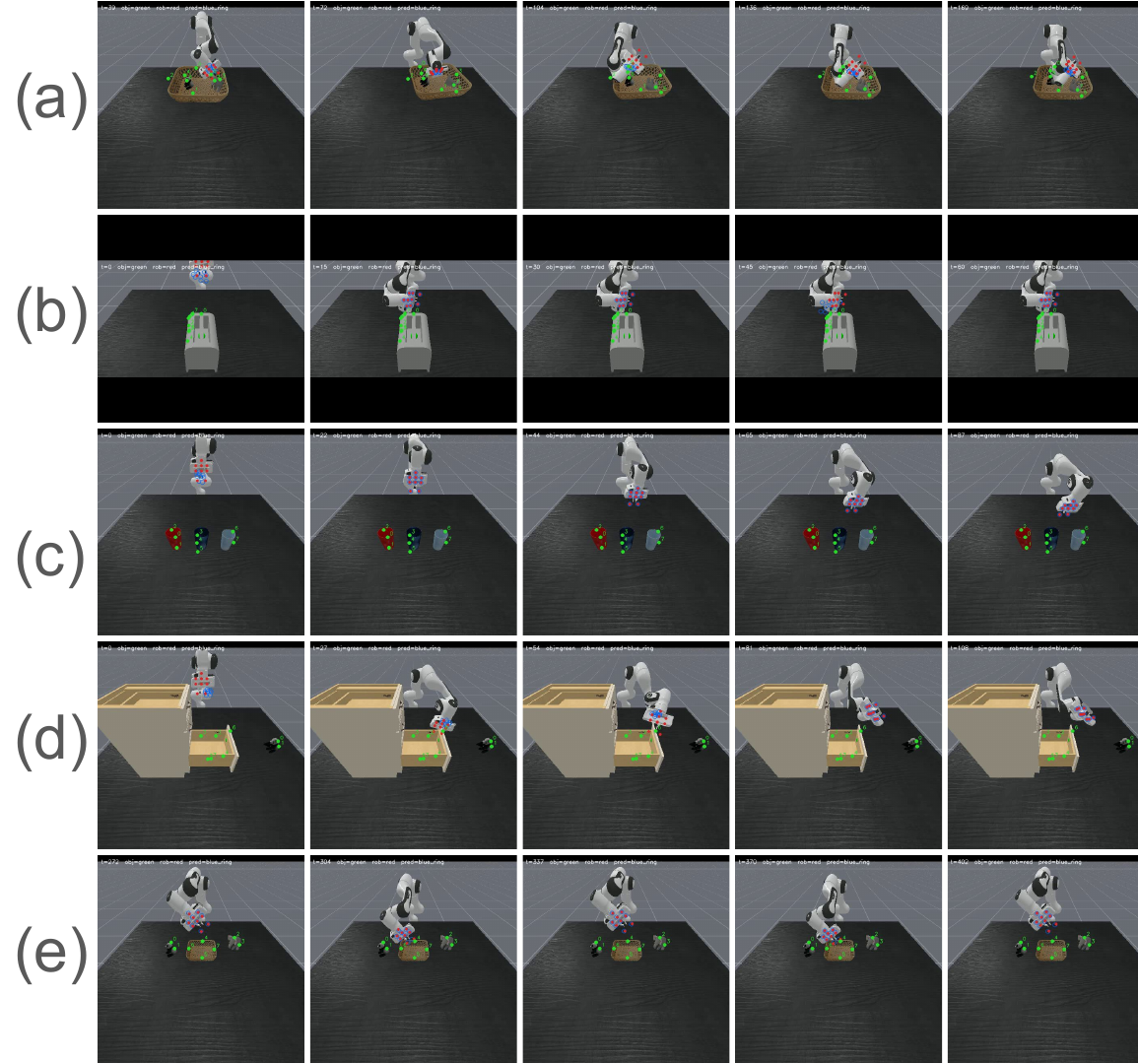}
    \caption{Qualitative rollouts for the extended tasks: (a) empty basket, (b) press a toaster button, (c) stack three cups, (d) place a toy in a drawer and close the drawer, and (e) place toys in a basket. In all tasks the policy follows the overall trajectory, moving toward the cups, toys, or button, but fails to execute the precise manipulation required to press the button or grasp the target object.}
    \label{fig:point_policy_rollout}
\end{figure}

Second, Point Policy operates on 3D keypoint coordinates and learns geometric relationships between channels rather than visual features. Although our sim-side and training-side labels follow the same ordering, the underlying keypoints do not correspond to exactly the same physical features. Small drift in where each channel is anchored (a few millimeters to a couple of centimeters between train and sim) shifts the relational offsets the policy learned. Combined with the approximately 2–3 cm precision floor of the rigid-transform action recovery, visible as the frequent reflection corrections (det(R) $<$ 0) during rollout, the policy can consistently navigate to the correct sub-region of the workspace but cannot reliably close the last few centimeters required to grasp objects whose graspable surfaces are themselves only a few centimeters wide as illustrated in Figure~\ref{fig:point_policy_rollout}. This is exactly why the previously evaluated non-prehensile tasks (push cube, close drawer) achieved markedly higher success rates: their action requirements, translating the object or pushing a handle, are far more tolerant of these few-centimeter errors than a precision grasp. Our extended tasks are predominantly prehensile, so the same precision floor that was harmless before now compounds with the keypoint-identity drift to make contact with the graspable surface unreliable.

% \begin{table}[t]
% \centering
% \scriptsize
% \begin{tabular}[width=0.48\textwidth]{llcc}
% \toprule
% \textbf{Model} & \textbf{Task} & \textbf{Planner Data} & \textbf{Teleop Data} \\
% \midrule
% \multirow{3}{*}{Vid. $\pi_{0.5}$-BPP}
%     & Bowl  & $\mathbf{1.00 \pm 0.00}$ & $0.87 \pm 0.09$ \\
%     & Drawer    & $1.00 \pm 0.00$ & $0.97 \pm 0.03$ \\
%     & Toy   & $0.53 \pm 0.09$ & $\mathbf{0.73 \pm 0.03}$ \\
% \midrule
% \multirow{3}{*}{UniSkill}
%     & Bowl  & $0.03 \pm 0.03$ & $0.10 \pm 0.06$ \\
%     & Drawer    & $1.00 \pm 0.00$ & $1.00 \pm 0.00$ \\
%     & Toy & $0.00 \pm 0.00$ & $0.00 \pm 0.00$ \\
% \bottomrule
% \end{tabular}
% \caption{Success rates of policies trained using planner-generated versus teleoperated robot trajectories. Bold indicates the higher-performing training source for each model-task pair.\vspace{-6mm}}
% \label{tab:planner_vs_teleop}
% \end{table}

\subsection{Comparison for Planner-Generated Trajectories and Teleoperated Trajectories}
RoboReel uses planner-generated robot trajectories for training to enable scalable and standardized evaluation across tasks and methods.
To examine whether model performance when trained with planner-generated trajectories reflects the performance when trained with human-teleoperated trajectories, we conduct an additional comparison using both data sources. 
We collect 50 planner-generated and 50 expert-teleoperated robot trajectories for three representative tasks and train video-conditioned $\pi_{0.5}$ and UniSkill separately on each dataset. We evaluate each trained policy over three runs of $10$ rollouts using the same held-out human demonstrations. 
We do not expect the two training data sources to produce identical performance, but instead examine whether relative performance trends across models and tasks are preserved.

The results are presented in Table~\ref{tab:planner_vs_teleop}. 
Despite differences in absolute success rates between the two training conditions, we observe consistent model-task performance trends, with a Pearson correlation of $r=0.96$ and a Spearman rank correlation of $\rho=0.92$. 
Although planner-generated and teleoperated trajectories do not necessarily follow the same distribution, these results suggest that training with RoboReel's planner-generated data can reflect the relative performance ranking observed when training with teleoperated data.

\section{Model Architecture for Video-Conditioned $\pi_{0.5}$}
\label{appendix:arch:vapi05}

Our video-augmented Pi-0.5 policy builds on the perceptual-memory and
memory-as-context design introduced by \citet{robomme}, but repurposes
the memory tokens for learning from a human demonstration video rather than
from the robot's own past observations. Specifically, the policy is built on
top of the public \texttt{openpi/pi05\_base} checkpoint~\citep{pi05}
and preserves its SigLIP-So400m/14 image tower, Gemma VLM expert, and
flow-matching action expert. The only added component is a lightweight
demonstration encoder that converts a human video into a fixed-length visual
prefix. This prefix is prepended to the standard Pi-0.5 input sequence and is
processed jointly with the robot observation and language tokens.

\subsection{Video-Conditioned Policy Formulation}
\label{appendix:arch:vapi05:formulation}

We formulate demonstration-conditioned manipulation as action prediction from
three sources of information: a human demonstration video, the robot's current
multi-view observation history, and a generic language prompt. Let
\[
    D = \{I^{h}_{1}, I^{h}_{2}, \ldots, I^{h}_{T_h}\}
\]
denote a human demonstration video for the target task, where each
$I^{h}_{t}$ is an RGB frame. Let
\[
    O_t = \{I^{r,v}_{t-k:t}\}_{v=1}^{V}
\]
denote the robot observation history at control step $t$, consisting of
multi-view RGB observations over an observation horizon of length $k{+}1$.
The policy predicts an action chunk
\[
    A_t = (a_t, a_{t+1}, \ldots, a_{t+H-1})
\]
conditioned on the demonstration and the current robot observations:
\[
    \pi_\theta(A_t \mid D, O_t, \ell),
\]
where $\ell$ is a fixed task prompt. In our experiments, we set
$\ell=$``\texttt{complete the task}'' for all tasks, so the human
demonstration is the only task-specific conditioning signal.

The demonstration video is first mapped to a fixed-length sequence of
demonstration tokens:
\[
    Z_D = g_\phi(D) \in \mathbb{R}^{B \times d},
\]
where $B{=}512$ is the demonstration-token budget and $d{=}2048$ is the
Pi-0.5 token dimension. The robot observations and language prompt are
embedded by the original Pi-0.5 encoders into token sequences
$Z_O$ and $Z_\ell$. The VLM expert receives the concatenated prefix
\[
    Z_{\mathrm{prefix}} = [Z_D; Z_O; Z_\ell],
\]
and the action expert predicts the velocity field used by the flow-matching
objective:
\[
    v_\theta(x_t, t \mid Z_{\mathrm{prefix}}),
\]
where $x_t$ is the noisy action chunk at denoising time $t$. Training follows
the original Pi-0.5 conditional flow-matching objective, with the
demonstration tokens acting as additional visual context.

\subsection{Base Pi-0.5 Components}
\label{appendix:arch:vapi05:basic}

\paragraph{Input tokenization.}
The policy receives three types of tokens: demonstration tokens, robot
observation tokens, and language tokens. Robot observations are encoded from
the \texttt{base\_camera} and \texttt{hand\_camera} RGB streams. Language
tokens encode the fixed prompt \texttt{``complete the task''}. All tokens are
projected to the Pi-0.5 hidden dimension $d{=}2048$. Since the language
prompt is identical across tasks, task identity must be inferred from the
demonstration prefix.

\paragraph{VLM and action experts.}
We preserve the two-expert structure of Pi-0.5. The VLM expert fuses the
visual and language prefix tokens, while the action expert predicts robot
action chunks conditioned on the VLM features and the denoising timestep. We
initialize both experts from \texttt{pi05\_base} and fine-tune them jointly
with the demonstration encoder.

\paragraph{Attention structure.}
The demonstration, robot-image, and language tokens form the prefix sequence.
Attention is bidirectional within this prefix, allowing the model to jointly
reason over the human video, the current robot scene, and the prompt. The
action suffix contains the noisy action tokens and uses the standard
prefix-LM structure: action tokens attend to the prefix and causally attend
within the action sequence. We implement this token-level masking with the
\texttt{ar\_mask} and \texttt{na\_mask} interface in the \texttt{LfOPi0}
forward pass.

\paragraph{Flow-matching action prediction.}
The action expert predicts action chunks using conditional flow matching. A
velocity field is trained to map Gaussian noise to the target action chunk
conditioned on the prefix representation. The denoising timestep is sampled
per batch element as $t\sim\mathrm{Beta}(1.5,1)$ and scaled to the interval
$(0.001,1)$, matching the Pi-0.5 training setup.

\subsection{Human Demonstration Encoder}
\label{appendix:arch:vapi05:perceptual}

The human demonstration encoder converts a single front-view human video into
a fixed-length visual prefix. The prefix has a budget of $B{=}512$ tokens and
is constructed from selected frames of the demonstration video.

\paragraph{Offline visual feature extraction.}
For efficiency, demonstration features are precomputed and cached before
policy training. Each $224{\times}224$ front-view demonstration frame is
passed through the SigLIP-So400m/14 image tower from the
\texttt{pi05\_base} checkpoint. The resulting patch features are
average-pooled to a $4{\times}4$ spatial grid, yielding $P{=}16$ visual
tokens per frame. Each token has dimension $2048$. We also cache
$8{\times}8$ and $2{\times}2$ pooled grids, but the deployed configuration
uses only the $4{\times}4$ grid. The SigLIP image tower is frozen during this
precomputation step.

\paragraph{Spatiotemporal positional encoding.}
Each demonstration token is augmented with a 3-D sinusoidal positional
embedding that encodes its spatial and temporal location. The embedding has
dimension $768$ and is divided across the $y$, $x$, and $t$ axes using
sinusoidal sine-cosine features. Spatial coordinates are computed with
respect to the original $16{\times}16$ patch grid and then mapped to the
pooled $4{\times}4$ grid so that each pooled token is positioned at the
center of its corresponding patch group. Temporal coordinates are indexed by
the absolute frame number in the demonstration video. Spatial frequencies use
base $1{,}000$, and temporal frequencies use base $10{,}000$.

\paragraph{Uniform frame sampling.}
In the uniform variant, we sample $T_{\mathrm{frame}}{=}32$ frames evenly
across the demonstration video. Since each frame contributes $P{=}16$ visual
tokens and we use a single human-video view, the resulting prefix length is
\[
    T_{\mathrm{frame}} \cdot P \cdot V
    = 32 \cdot 16 \cdot 1
    = 512,
\]
which exactly matches the demonstration-token budget.

\paragraph{BPP keyframe sampling.}
We also evaluate a keyframe-aware sampling variant, BPP~\cite{bpp}. Given a set of
task-relevant keyframes, all annotated keyframes are included whenever they
fit within the frame budget. The remaining frame slots are sampled from the
non-keyframe frames with probability inversely related to their distance from
the nearest keyframe. Let $\mathcal{K}$ denote the set of selected keyframes
and let
\[
    d_{\mathrm{kf}}(i) = \min_{j \in \mathcal{K}} |i-j|
\]
be the temporal distance from frame $i$ to its closest keyframe. Non-keyframe
frames are sampled with weights proportional to
\[
    w_i \propto \frac{1}{d_{\mathrm{kf}}(i) + \epsilon},
\]
where $\epsilon$ avoids division by zero. Thus, frames closer to annotated
keyframes are more likely to be selected. The uniform and BPP variants differ
only in the frame sampling rule; all other architectural components are
shared.

\paragraph{Token construction.}
For each selected demonstration frame, the pooled SigLIP feature and the
3-D positional embedding are combined into a Pi-0.5-compatible token. Let
$f_{i,p}\in\mathbb{R}^{2048}$ be the SigLIP feature for patch $p$ in selected
frame $i$, and let $e_{i,p}\in\mathbb{R}^{768}$ be its spatiotemporal
positional embedding. The encoder computes
\[
    z_{i,p}
    =
    W_{\mathrm{joint}}
    \left[
        f_{i,p};
        \mathrm{silu}(W_{\mathrm{pos}} e_{i,p})
    \right],
\]
where $W_{\mathrm{pos}}:\mathbb{R}^{768}\rightarrow\mathbb{R}^{768}$ and
$W_{\mathrm{joint}}:\mathbb{R}^{2816}\rightarrow\mathbb{R}^{2048}$ are
learned linear projections. State-embedding fusion is disabled in our
configuration. The resulting sequence of tokens
$\{z_{i,p}\}_{i,p}$ forms the demonstration prefix $Z_D$.

The demonstration encoder consists only of these two linear layers and adds
approximately 6.5M parameters. It does not introduce additional transformer
blocks, attention heads, or expert branches.

\subsection{Demonstration-as-Context Integration}
\label{appendix:arch:vapi05:ctx}

The demonstration prefix is integrated by prepending it to the standard
Pi-0.5 prefix sequence. Concretely, the VLM expert receives
\[
    [Z_D; Z_O; Z_\ell],
\]
where $Z_D$ contains the human demonstration tokens, $Z_O$ contains the
current robot observation tokens, and $Z_\ell$ contains the fixed language
prompt tokens. These tokens are processed together by the PaliGemma stack
under the prefix attention mask. The action expert then predicts the
flow-matching velocity field conditioned on the resulting prefix features.

This integration mechanism treats the human demonstration as visual context
for the current control problem. The model can therefore compare the human
video with the current robot observation before predicting the next action
chunk. Importantly, the action head, action horizon, denoising procedure, and
flow-matching objective are unchanged from Pi-0.5.

\paragraph{Summary.}
Video-Augmented Pi-0.5 is Pi-0.5 with a length-$512$ visual demonstration
prefix. The prefix encodes a $32$-frame human demonstration video using
frozen SigLIP features, spatiotemporal positional embeddings, and a small
linear projection module. Apart from this demonstration encoder and the
prepended visual prefix, the underlying Pi-0.5 architecture is left
unchanged.

\section{Training Details}
\label{appendix:training_details}

Each model is trained twice with identical hyperparameters: once on the
no-distraction training set and once on the with-distraction training set
(see the main paper). Unless otherwise specified, all runs use a single
NVIDIA H200 GPU with 144~GB memory and seed 42. We save checkpoints every
10k steps and report results from the final checkpoint. The final checkpoint
is at 50k steps for all variants except the Lerobot Pi-0.5 FFT
language-conditioned baseline, which is trained for 30k steps.

% \subsection{BC-Z}
% \label{appendix:training:bcz}

% We train the canonical BC-Z architecture from~\cite{bcz} using the authors'
% released codebase. The image-to-waypoint network is a ResNet-50 with
% FiLM-conditioning on the demonstration-video embedding. The video encoder is
% a frame-wise ResNet-18 applied to $T{=}20$ frames, producing a 512-d clip
% embedding. The auxiliary text-cosine alignment loss uses the Universal
% Sentence Encoder (USE-Large) as a frozen target encoder. Input images are
% cropped to $100{\times}100$, and the model predicts $H{=}10$ future
% waypoints. We use a cosine-alignment loss weight of
% $\lambda_{\text{cos}}{=}1.0$. Optimization uses TensorFlow Adam with learning
% rate $3{\times}10^{-4}$, no warmup or decay, batch size 32, and 50,000
% gradient steps. Sentence embeddings for the text-alignment loss are
% precomputed and cached in \texttt{bcz\_sentence\_embeddings.npz}.

\subsection{Vid2Robot}
\label{appendix:training:vid2robot}

We implement Vid2Robot following the description in the original
paper~\cite{vid2robot}. The visual backbone is a pretrained ViT-B/16 at
$224{\times}224$ resolution. The prompt and state streams each pass through a
2-layer Perceiver Resampler with 64 latents, 12 heads, and head dimension 64.
The State-Prompt Encoder and Action Decoder are 4-layer Transformers with
hidden dimension 768 and 8 attention heads. The model receives
$T_{\text{prompt}}{=}16$ prompt frames and $T_{\text{state}}{=}8$ state
frames. The action head predicts a chunk of $C{=}4$ actions, with each action
dimension discretized into 256 bins for the 8-D \texttt{pd\_joint\_pos}
action space.

The training objective is the mean of four losses: (i) cross-entropy over
action tokens, (ii) Temporal Cycle Consistency between prompt and robot frame
projections, (iii) prompt--robot video-video SigLIP loss, and (iv) video-text
SigLIP loss against a SigLIP text-tower embedding of the task string.
Optimization uses AdamW with peak learning rate $3{\times}10^{-4}$, final
learning rate $1{\times}10^{-6}$, weight decay 0.05, linear warmup for 2,000
steps, and cosine decay to the final learning rate. We train with batch size
32 for 50,000 steps. To fit training on a single H200, we apply per-block
gradient checkpointing to the ViT.

\subsection{UniSkill}
\label{appendix:training:uniskill}

We use the public UniSkill Inverse-Skill-Dynamics (ISD) encoder off the
shelf, without fine-tuning on our data. The encoder weights are loaded from
the authors' released pretrained checkpoint. This follows the protocol of
the original paper, where the pretrained ISD checkpoint is not fine-tuned on
the target human data. We run the encoder once over every robot frame and
every human demonstration in the LfO data to precompute a 64-d skill-goal
embedding for each clip. Only the downstream policy is trained.

\textbf{Downstream diffusion policy.}
We train a Robomimic diffusion policy conditioned on the precomputed
skill-goal embedding. The visual encoder consists of two ResNet-18
SpatialSoftmax heads, each with 32 keypoints, applied to the front-view and
wrist-camera observations. We use random $116{\times}116$ crops. Low-dimensional
inputs include gripper and joint states. The denoiser is a conditional 1-D
UNet with channels $[256, 512, 1024]$, kernel size 5, 8 groups, and
diffusion-step embedding dimension 256. The model is trained with DDPM-style
noise prediction using 100 training timesteps and evaluated with DDIM using
10 inference steps. We use a squared-cosine-cap-$v_2$ noise schedule with
\texttt{clip\_sample=True}. The observation, action, and prediction horizons
are 2, 8, and 16, respectively.

EMA is enabled with power 0.75. Optimization uses Adam with learning rate
$1{\times}10^{-4}$, no $L_2$ regularization, batch size 256, and 1,000 epochs
of 100 iterations each, for a total of 100,000 gradient steps. We use seed 0.
Actions are normalized per dimension using min-max statistics. We use the
\texttt{skill\_aug} setting with \texttt{aug\_num=5}, where the skill
embedding is replaced by a same-task neighbor with probability determined by
the augmentation count.

\subsection{Video-Conditioned $\pi_{0.5}$ (Uniform Frame Sampling)}
\label{appendix:training:pi05_uniform}

The video-conditioned Pi-0.5 variant fine-tunes the public Pi-0.5 base
checkpoint with our demonstration adapter. We represent the human
demonstration using a token budget of 512 and a frame budget of 32, with 16
tokens per image. The adapter uses an observation horizon of 16, mean
pooling, a 2048-d demonstration-token dimension, enabled positional
embeddings, and disabled state embeddings. Demonstration frames are sampled
uniformly across the human video. Robot observations use both the front-view
and wrist cameras, while human demonstrations are collected from the
front-view camera.

Optimization uses JAX FSDP on one device with AdamW
($\beta_1{=}0.9$, $\beta_2{=}0.95$, $\epsilon{=}10^{-8}$, weight decay
$10^{-10}$), gradient-norm clipping at 1.0, and a cosine learning-rate
schedule with peak learning rate $5{\times}10^{-5}$. The schedule uses 1,000
warmup steps and decays to $0.1{\times}$ the peak learning rate. EMA is
applied to the weights with decay 0.999. We train with batch size 64 for
50,000 steps using seed 42. The text prompt is fixed to
``\textit{complete the task}'', so the model receives no language information
beyond the demonstration video.
\begin{figure}[t]
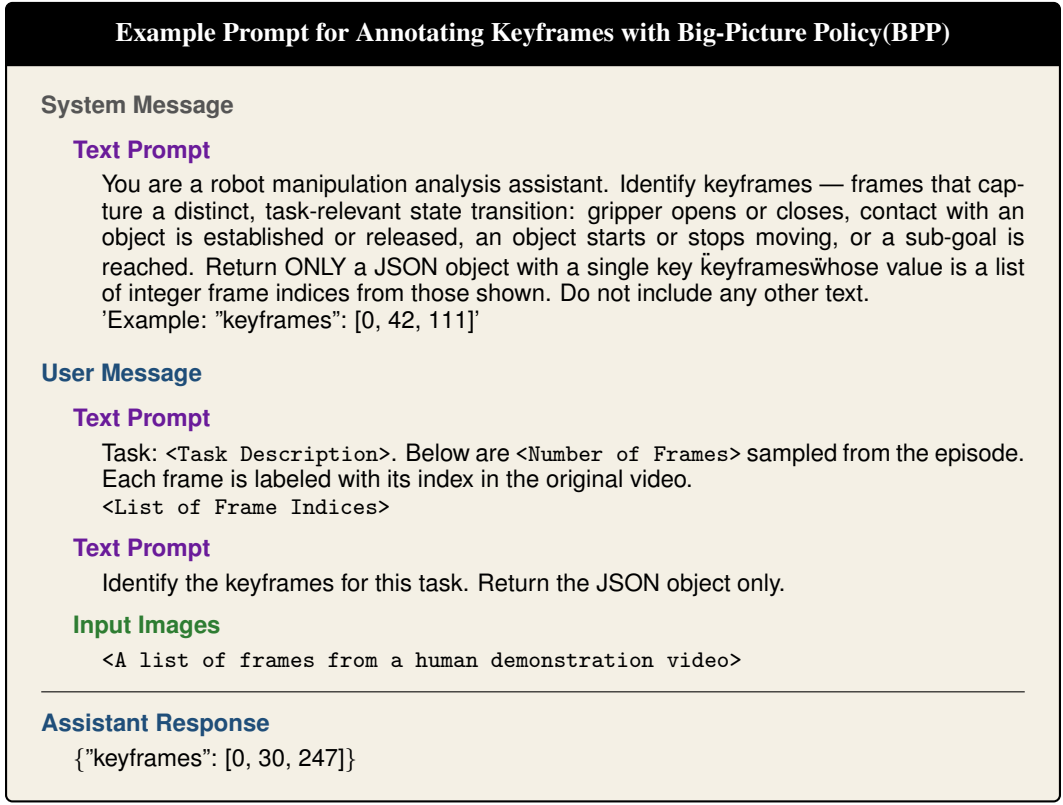

\centering
\begin{tcolorbox}[
  enhanced,
  breakable,
  colback=PromptBack,
  colframe=black,
  boxrule=0.8pt,
  arc=2pt,
  left=10pt,right=10pt,top=8pt,bottom=8pt,
  title={Example Prompt for Annotating Keyframes with Big-Picture Policy(BPP)},
  coltitle=white,
  colbacktitle=black,
  fonttitle=\bfseries,
  center title,
  toptitle=4pt,
  bottomtitle=4pt,
]

{\sffamily
\promptsection{PromptGray}{System Message}
\vspace{4pt}
\promptsubsection{PromptPurple}{Text Prompt}
\begin{promptindent}
    \begin{promptindent}
        You are a robot manipulation analysis assistant. Identify keyframes — frames that capture a distinct, task-relevant state transition: gripper opens or closes, contact with an object is established or released, an object starts or stops moving, or a sub-goal is reached. Return ONLY a JSON object with a single key \"keyframes\" whose value is a list of integer frame indices from those shown. Do not include any other text. \\
        'Example: {"keyframes": [0, 42, 111]}'
    \end{promptindent}
\end{promptindent}

\vspace{8pt}
\promptsection{PromptBlue}{User Message}
\vspace{4pt}
\promptsubsection{PromptPurple}{Text Prompt}
\begin{promptindent}
\begin{promptindent}
    Task: \texttt{<Task Description>}. Below are \texttt{<Number of Frames>} sampled from the episode. Each frame is labeled with its index in the original video.\\
    \texttt{<List of Frame Indices>}
\end{promptindent}
\end{promptindent}
\vspace{4pt}
\promptsubsection{PromptPurple}{Text Prompt}
\begin{promptindent}
\begin{promptindent}
    Identify the keyframes for this task. Return the JSON object only.
\end{promptindent}
\end{promptindent}
\vspace{4pt}
\promptsubsection{PromptGreen}{Input Images}
\begin{promptindent}
\begin{promptindent}
    \texttt{<A list of frames from a human demonstration video>}
\end{promptindent}
\end{promptindent}

\vspace{8pt}
\hrule
\vspace{8pt}

\promptsection{PromptBlue}{Assistant Response}

\begin{promptindent}
    \{"keyframes": [0, 30, 247]\}
\end{promptindent}

}

\end{tcolorbox}
\caption{Example prompt for generating keyframe annotations using the VLM-based keyframe annotation framework. This framework is inspired by BPP, and we use GPT-5.4-mini to as the backbone VLM. The downstream video-conditioned $\pi_{0.5}$ VLA policy select frames from the videos using these keyframe annotations.}
\label{fig:keyframe_prompt}
\end{figure}

\begin{figure}[t]
\centering
\begin{tcolorbox}[
  enhanced,
  breakable,
  colback=PromptBack,
  colframe=black,
  boxrule=0.8pt,
  arc=2pt,
  left=10pt,right=10pt,top=8pt,bottom=8pt,
  title={Example Prompt for Using VLM to Annotate Human Videos},
  coltitle=white,
  colbacktitle=black,
  fonttitle=\bfseries,
  center title,
  toptitle=4pt,
  bottomtitle=4pt,
]

{\sffamily

\promptsection{PromptBlue}{User Message}

\promptsubsection{PromptGreen}{Input Images}
\begin{promptindent}
\begin{promptindent}
    \texttt{<A list of frames from a human demonstration video>}

\end{promptindent}
\end{promptindent}

\vspace{4pt}

\promptsubsection{PromptPurple}{Text Prompt}
\begin{promptindent}
\begin{promptindent}
    These images are frames sampled in chronological order from a video of a person
performing a manipulation task. Write the complete and clean task instruction
for this video so a different human operator can reproduce the task without
looking at the video.
\end{promptindent}
\end{promptindent}

\vspace{8pt}
\hrule
\vspace{8pt}

\promptsection{PromptBlue}{Assistant Response}

\begin{promptindent}
    Place the panda-shaped object into the bottom drawer of the white mesh organizer.
\end{promptindent}

}

\end{tcolorbox}
\caption{Example prompt for using VLM to generate language annotation for human demonstration videos. The generated language instructions are consumed by the downstream $\pi_{0.5}$ VLA policy.}
\label{fig:annotation_prompt}
\end{figure}

\subsection{Video-Conditioned $\pi_{0.5}$ (BPP keyframe sampling)}
\label{appendix:training:pi05_bpp}
The BPP-$\pi_{0.5}$ variant uses the same architecture and optimization
settings as the uniform frame-sampling variant. The only difference is the
human-demonstration frame sampling strategy. 
For this variant, we use a keyframe annotation framework inspired by BPP to
precompute keyframe annotations and assign each frame a sampling weight based
on its distance to the closest selected keyframe. Frames closer to selected
keyframes receive higher sampling weights.

Fig.~\ref{fig:keyframe_prompt} describes the prompt we use to annotate keyframes with a self-implemented VLM framework inspired by BPP framework~\cite{bpp}. 
We use GPT-5.4-mini~\cite{gpt5} as the backbone VLM to select keyframes for input human demonstration videos.
The original BPP framework use all frames of the videos as inputs for keyframe annotations. To reduce the computational cost, we downsample the human videos to $2$-fps uniformly, and provide the list of indices for the selected frames for the VLM. The VLM then returns the selected keyframes from the list of candidate frames in the json format.

\subsection{Language-conditioned $\pi_{0.5}$(VLM Upstream)}
\label{appendix:training:lerobot_pi05_fft}

This baseline is the language-conditioned $\pi_{0.5}$
variant, and does not take in video inputs. 
We fully fine-tune language-conditioned $\pi_{0.5}$ using the lerobot implementation.
Training uses robot trajectories with base and wrist cameras at
$224{\times}224$ resolution, robot state, and action chunks. We finetune the action expert of the $\pi_{0.5}$ model and keep the VLM expert frozen during training. The language instruction for each robot sample is generated by Qwen3-VL-4B from the corresponding human
demonstration of the same task. Fig.~\ref{fig:annotation_prompt} shows the prompt we use to invoke the VLM, and an example response from the VLM model. We use the huggingface checkpoint for video annotations.

Optimization uses PyTorch AdamW with peak learning rate
$2.5{\times}10^{-5}$ and cosine decay to $0.1{\times}$ the peak learning rate using cosine annealing LR over the full training schedule, without a separate warmup phase. The optimizer $\beta$ values, $\epsilon$, and weight decay are inherited from the lerobot preset. Gradients are clipped using the threshold specified by the configuration. 
We train with batch size 32, bfloat16 precision, gradient checkpointing, 30,000 gradient steps. 
Input normalization statistics are estimated from 5,000 randomly sampled training batches before training begins. We store checkpoints at every $10,000$ steps and evaluate the last checkpoint.

\subsection{PointPolicy}
We follow Point-Policy ~\cite{pointpolicy} to train an imitation policy directly from human demonstrations.
We extract keypoint trajectories from the human videos and retarget them into the robot end-effector trajectories in the calibrated environment. 
These end-effector keypoints, together with the task-specific object keypoints, are used to train the policy. Given the current end-effector keypoints and object keypoints, the model predicts the next end-effector keypoints. The details of the keypoint annotation and the mapping to simulation of  the robot actions are described in Appendix~\ref{appendix:benchmark_detailed_descriptions}.

The PointPolicy baseline is trained independently for each task. We train the policy for $90,000$ gradient steps with a batch size of $128$, using AdamW optimizer.
Predictions are executed using a waypoint-based trajectory controller. Each prediction defines the next target waypoint, and the controller drives the end-effector toward it until the waypoint is reached or a per-waypoint timeout elapses. The policy is then queried again to generate the next waypoint. This process continues until the task is completed or the maximum episode length is reached.
The PointPolicy is evaluated in the simulated environments under two settings, with distraction objects and without distraction objects. We use the same set of distraction objects for the PD environment as the WD and ED environments. The details are described in Appendix~\ref{appendix:benchmark_details}.

\begin{table*}[t]
\centering
\small
\resizebox{\textwidth}{!}{%
\begin{tabular}{c|cccccc}
\toprule

\multirow{3}{*}{\makecell[c]{\textbf{Dataset}\\\textbf{Name}}} 
& \multirow{3}{*}{\makecell[c]{\textbf{Total}\\\textbf{\#Tasks}}}
& \multirow{3}{*}{\makecell[c]{\textbf{Real-world}\\\textbf{Human Demo.}\\\textbf{\#Hours}}}
& \multirow{3}{*}{\makecell[c]{\textbf{Total}\\\textbf{\#Human}\\\textbf{ Demo.}}}
& \multirow{3}{*}{\makecell[c]{\textbf{Avg. \#Secs}\\\textbf{Human Demo.}}}
& \multirow{3}{*}{\makecell[c]{\textbf{Total}\\\textbf{\#Robot Demo.}}}
& \multirow{3}{*}{\makecell[c]{\textbf{Avg. \#Steps}\\\textbf{Robot Demo.}}}\\
& & & & & & \\
& & & & & & \\
\midrule
RoboCerebra~\cite{robocerebra}  & $1,060$ & N/A & N/A     &  N/A  & $1,000$ & $2,972$ \\
LIBERO~\cite{libero}            & $130$   & N/A & N/A     &  N/A  & $6,500$ & $162$  \\
RoboCasa~\cite{robocasa}        & $100$   & N/A & N/A     &  N/A  & $100$K+ & -    \\
VLABench~\cite{vlabench}        & $100$   & N/A & N/A     &  N/A  & $1,600$ & $115$ \\
MimicGen~\cite{mimicgen}        & $12$    & N/A & N/A     &  N/A  &  $48$K+ & - \\
\midrule
EgoVerse~\cite{egoverse}        & $1,965$ & $1,362$ & $79,620$&   -   &  N/A    &  N/A \\
Human2Robot~\cite{human2robot}  & $8$     &  -      & $2,600$ &   -   & $2,600$ &   -  \\
RH20T~\cite{rh20t}              & $147$   & -       & $110$K+ &   -   & $110$K+ &   -  \\
MimicDroid~\cite{mimicdroid}    & $12$    & N/A     & N/A     &  N/A  & N/A     &   -  \\
RoboReel                        & $10$    & $9$     & $3,000$ & $11.56$ & $2,000$ & $304$\\
\bottomrule
\end{tabular}%
}
\caption{\textbf{Overall statistics comparison between RoboReel and other existing benchmarks.} Compared to existing datasets and benchmarks, RoboReel features paired real-world human demonstration videos and portable and scalable evaluation environment. This allows a fair comparison for learning-from-observation frameworks using our benchmark.}
\label{tab:benchmark_statistics_comparison}
    
\end{table*}

\begin{table*}[t]
\centering
\small
\resizebox{\textwidth}{!}{%
\begin{tabular}{l rrrr rr rrr rr}
\toprule
& \multicolumn{6}{c}{\textbf{Human videos}} 
& \multicolumn{5}{c}{\textbf{Robot trajectories}} \\
\cmidrule(lr){2-7} \cmidrule(lr){8-12}
& \multicolumn{4}{c}{\textbf{Counts}}
& \multicolumn{2}{c}{\textbf{Duration (s)}}
& \multicolumn{3}{c}{\textbf{Counts}}
& \multicolumn{2}{c}{\textbf{Length (steps)}} \\
\cmidrule(lr){2-5} \cmidrule(lr){6-7}
\cmidrule(lr){8-10} \cmidrule(lr){11-12}
Task 
& Total & ND & WD & PD & Mean & Std
& Total & ND & WD & Mean & Std \\
\midrule
bowl\_in\_plate   & 300 & 100 & 100 & 100 & 12.68 & 3.97 & 200 & 100 & 100 & 262 & 19 \\
close\_drawer     & 300 & 100 & 100 & 100 &  7.46 & 1.55 & 200 & 100 & 100 & 185 & 22 \\
empty\_basket     & 300 & 100 & 100 & 100 & 14.01 & 4.43 & 200 & 100 & 100 & 498 & 57 \\
open\_lid         & 300 & 100 & 100 & 100 & 10.46 & 1.24 & 200 & 100 & 100 & 268 &  5 \\
pick\_cube        & 300 & 100 & 100 & 100 &  7.09 & 2.41 & 200 & 100 & 100 & 139 & 14 \\
press\_toaster    & 300 & 100 & 100 & 100 &  5.72 & 1.61 & 200 & 100 & 100 & 112 & 21 \\
push\_cube        & 300 & 100 & 100 & 100 &  5.73 & 0.94 & 200 & 100 & 100 & 135 &  6 \\
stack\_cups       & 300 & 100 & 100 & 100 & 17.84 & 1.79 & 200 & 100 & 100 & 515 & 50 \\
toys\_in\_basket  & 300 & 100 & 100 & 100 & 17.44 & 3.90 & 200 & 100 & 100 & 436 & 36 \\
toys\_in\_drawer  & 300 & 100 & 100 & 100 & 13.04 & 2.61 & 200 & 100 & 100 & 489 & 29 \\
\midrule
\textbf{Total / Avg.}
& \textbf{3000} & \textbf{1000} & \textbf{1000} & \textbf{1000} & 11.56 & ---
& \textbf{2000} & \textbf{1000} & \textbf{1000} & 304 & --- \\
\bottomrule
\end{tabular}%
}
\caption{
Per-task dataset statistics for human video demonstrations and robot trajectories. 
For each task, human videos include a total of $300$ demonstrations. Each of the ND, WD, and PD condition has $100$ demonstration videos. Robot trajectories include $200$ demonstrations, including $100$ trajectories for the ND and WD conditions each respectively. 
Human-video duration is reported in wall-clock seconds from the \texttt{front} camera at $30$~fps, and robot-trajectory length is reported as the number of discrete simulator control steps at $30$~Hz. 
The final row reports total counts and the average per-task duration or trajectory length.
}
\label{tab:combined_stats}
\end{table*}

\section{Details of the Benchmark}
\label{appendix:benchmark_details}
Table~\ref{tab:benchmark_statistics_comparison} shows the comparison of the statistics between RoboReel and existing datasets and benchmarks for robot learning and learning from human videos.
Compared to previous works, RoboReel provides paired real-world human videos and simulation trajectories for training, and paired task environments for fair and reproducible comparison for models from various model families on manipulation tasks of different categories.

\subsection{Dataset Statistics}
\label{appendix:dataset}

We report descriptive statistics for the benchmark in Table~\ref{tab:combined_stats}. 
This table provides detailed statistics for both human demonstration videos and robot trajectories. 
The statistics of the human videos are reported in seconds and the statistics of the robot trajectories are reported in the number of discrete time steps.

\paragraph{Human demonstration videos.}
Each task--condition pair includes 100 human demonstrations of the same
task performed by a human operator in front of a fixed table-top rig,
recorded simultaneously from \emph{four} cameras
(\texttt{egocentric}, \texttt{front}, \texttt{left}, \texttt{right}) at
$1280\times720$ resolution and $30$~fps. Across the 10 single-step tasks
the dataset totals $3{,}000$ demonstrations ($12{,}000$ mp4 files) spanning
roughly $9$~hours of clock time at the \texttt{front}-camera view (the
view our models consume). 

Six participants contribute demonstrations to the dataset, including five male and one female participant, with five right-handed and one left-handed participant. Demonstrators are instructed to perform the tasks at a controlled speed and minimize occlusions to maintain hand visibility. We do not require participants to dress uniformly or follow a fixed task strategy, allowing natural variations in appearance and task execution. Depending on the task configuration, participants also naturally use their non-dominant hand when necessary.

\paragraph{Robot trajectories.}
Each task--condition pair includes 100 robot trajectories collected
in ManiSkill 3 / SAPIEN with the same scene configuration and object
identity as the corresponding human demonstration pool. Every trajectory is
saved as one HDF5 holding the per-step robot state (\texttt{obs}, $36$-d),
8-d absolute \texttt{pd\_joint\_pos} actions, and the simulator's full
environment state, plus five synchronized $1280\times720$ RGB MP4s
(\texttt{front\_camera}, \texttt{hand\_camera}, \texttt{left\_camera},
\texttt{overhead\_camera}, \texttt{right\_camera}) at $30$~fps. Across the
10 single-step tasks the dataset totals $2{,}000$ trajectories.

\subsection{Additional Descriptions on the Benchmark}
\label{appendix:benchmark_detailed_descriptions}
\begin{table*}[t]
\centering
\small
\renewcommand{\arraystretch}{1.25}
\resizebox{\textwidth}{!}{%
\begin{tabular}{p{2.4cm} p{4.6cm} p{9.0cm}}
\toprule
\textbf{Task} & \textbf{Description} & \textbf{Success Criterion} \\
\midrule

\texttt{bowl\_in\_plate}
&
Pick up the bowl and place it on the plate.
&
The final position of the bowl is located on top of the plate. The xy-coordinate of the bowl is within the area of the plate, and the z-coordinate of the bowl must be on top of the plate.
\\
\midrule

\texttt{close\_drawer} 
&
Close the bottom drawer.
&
The drawer has been pushed inward by at least half of its open range.
\\
\midrule

\texttt{empty\_basket}$^\dagger$
&
Remove the two toys from the basket in any order.
&
Both toys are located outside the basket. Both toys must have final locations where either the xy-coordinates are outside the basket area, or the z-coordinates are beyond the height of the basket. 
\\
\midrule

\texttt{open\_lid}
&
Pick up the lid on the pot and place it on the table.
&
The lid is on the table and not on top of the pot. The xy-coordinate of the lid must be outside the pot area, and the z-coordinate of the lid must be between the table and the pot.
\\
\midrule

\texttt{pick\_cube}
&
Lift the cube above the table.
&
The z-coordinate of the cube is at least $3$cm above the table.
\\
\midrule

\texttt{press\_toaster}$^*$
&
Press the toaster down to the bottom.
&
The toaster's slider joint travels downward at least $80\%$ of its total range.
\\
\midrule

\texttt{push\_cube}
&
Push the cube away from its original position.
&
The cube has traveled for at least $5$cm initial position at any direction on the xy-plane.
\\
\midrule

\texttt{stack\_cups}$^{*\dagger}$
&
Stack the red cup on the dark blue cup, then stack the light blue cup
on top.
&
The three cups are arranged in the required vertical order: dark blue
at the bottom, red in the middle, light blue on top. All cups have the same xy-coordinate, and z-coordinates with a small distance following the order.
\\
\midrule

\texttt{toys\_in\_basket}$^\dagger$
&
Pick up the two toys and place them in the basket in any order.
&
Both toys are located inside the basket. Both toys must have final locations where the xy-coordinates are within the basket area, and the z-coordinates are below the height of the basket. 
\\
\midrule

\texttt{toys\_in\_drawer}$^\dagger$
&
Pick up the panda toy and place it in the drawer, then close the drawer.
&
The panda toy is located inside the open drawer, and the drawer is closed for at least half of its open range. The panda toy has a final position where the xy-coordinate is within the drawer area, and the z-coordinate is between the height of the lowest two drawers. The drawer has been pushed inward by at least half of its open range.
\\

\bottomrule
\end{tabular}%
}
\caption{\textbf{Detailed descriptions for each task.} We include the natural language task description for each task and the success criterion in the simulation. The $\dagger$ superscript denotes long-horizon task, and the $*$ superscript denotes high-precision tasks.}
\label{tab:task_descriptions}
\end{table*}

\begin{figure}
    \centering
    \includegraphics[width=\linewidth]{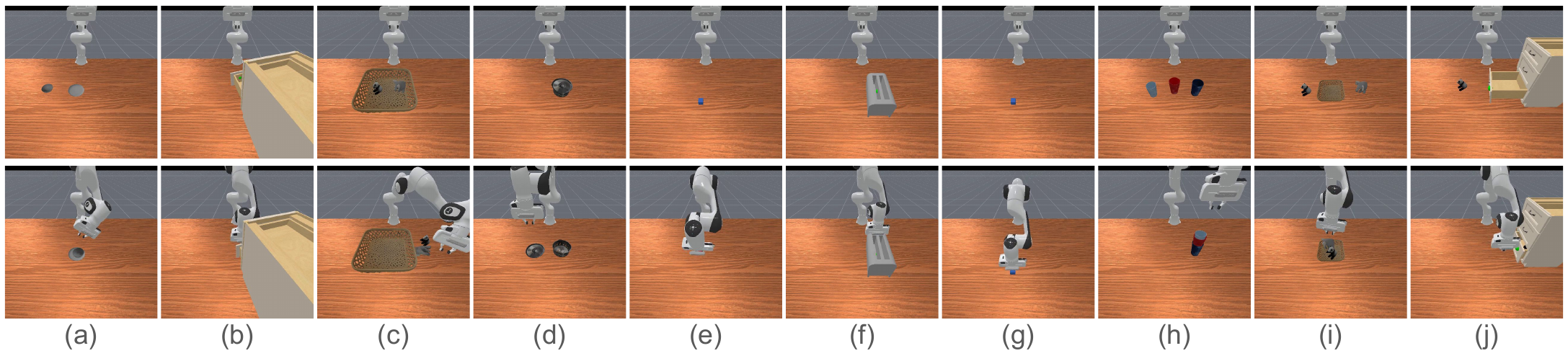}
    \caption{Examples of start and end configurations of a successful instance in the non-calibrated evaluation environment for each task-- (a) place a bowl in a plate, (b) close drawer, (c) empty basket, (d) open a lid, (e) pick a cube,(f) press a toaster button,  (g)push cube, (h) stack three cups, (i) place toys in a basket, (j) place a toy in a drawer and close the drawer. The top half of this figure shows the initial states and the bottom half shows the end states.}
    \label{fig:start_and_end_conditions}
\end{figure}

\begin{figure}
    \centering
    \includegraphics[width=\linewidth]{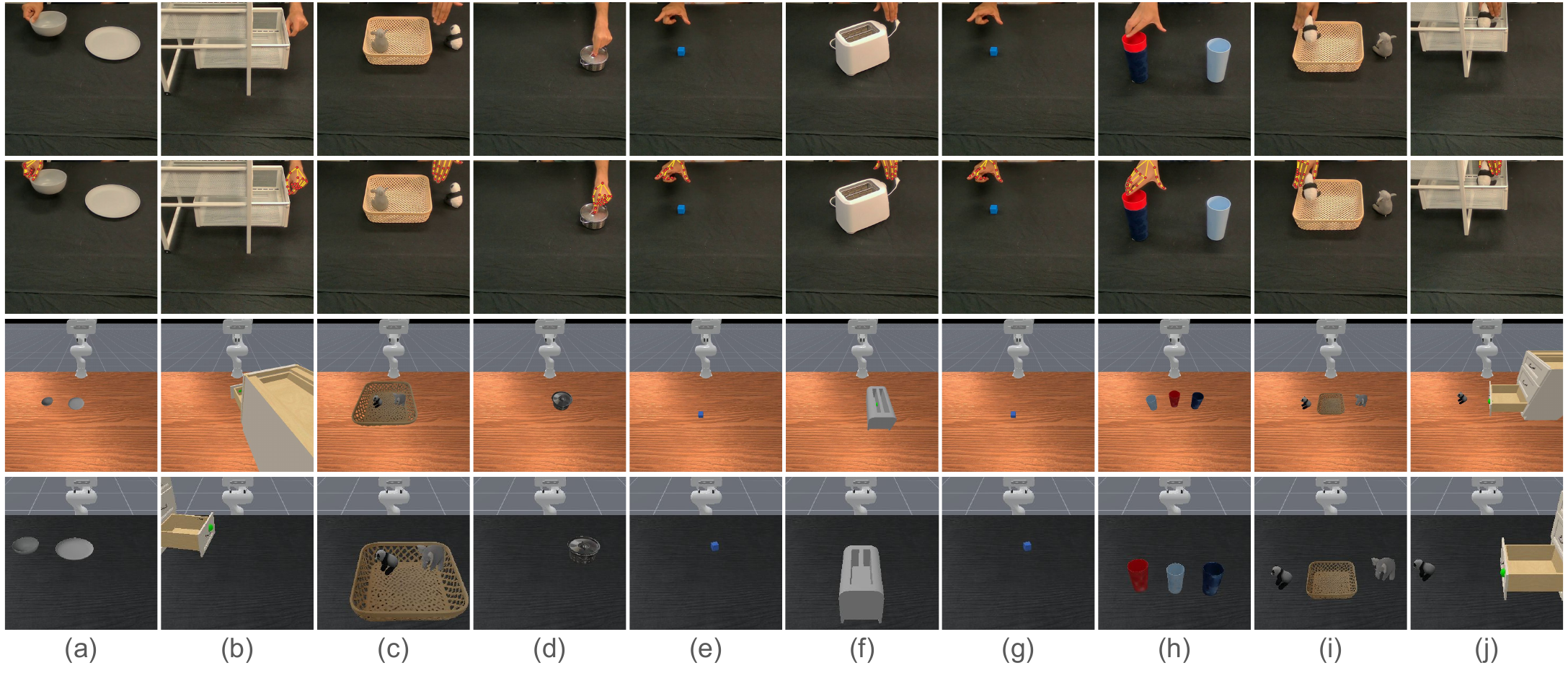}
    \caption{From top to bottom are frames from original human videos, human videos with keypoint annotations, non-calibrated task environment, and calibrated evaluation environments for each task -- (a) place a bowl in a plate, (b) close drawer, (c) empty basket, (d) open a lid, (e) pick a cube,(f) press a toaster button,  (g)push cube, (h) stack three cups, (i) place toys in a basket, (j) place a toy in a drawer and close the drawer. }
    \label{fig:corresponding_frames}
\end{figure}

\begin{figure}
    \centering
    \includegraphics[width=\linewidth]{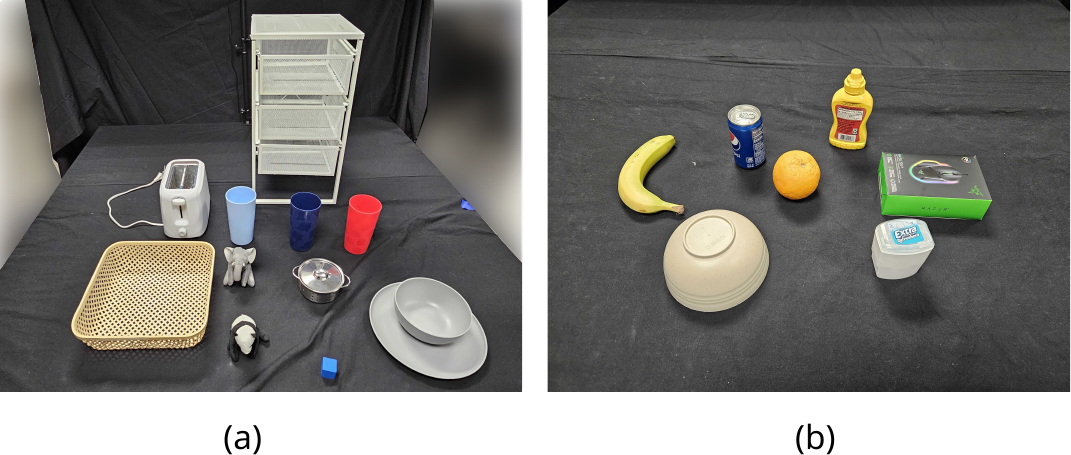}
    \caption{This figure shows the real-world objects we use. (a) The image of all the task relevant objects, including a toaster, a basket, a white drawer, three cups of different colors, an elephant soft toy, a pot and its lid, a panda soft toy, a blue cube, a gray bowl, and a gray plate. (b) The image of the real-world distraction objects, including a banana, a can of coke, a flipped bowl, an orange, yellow mustard, a mouse box, and a chewing gum box.}
    \label{fig:objects}
\end{figure}

\begin{figure}
    \centering
    \includegraphics[width=\linewidth]{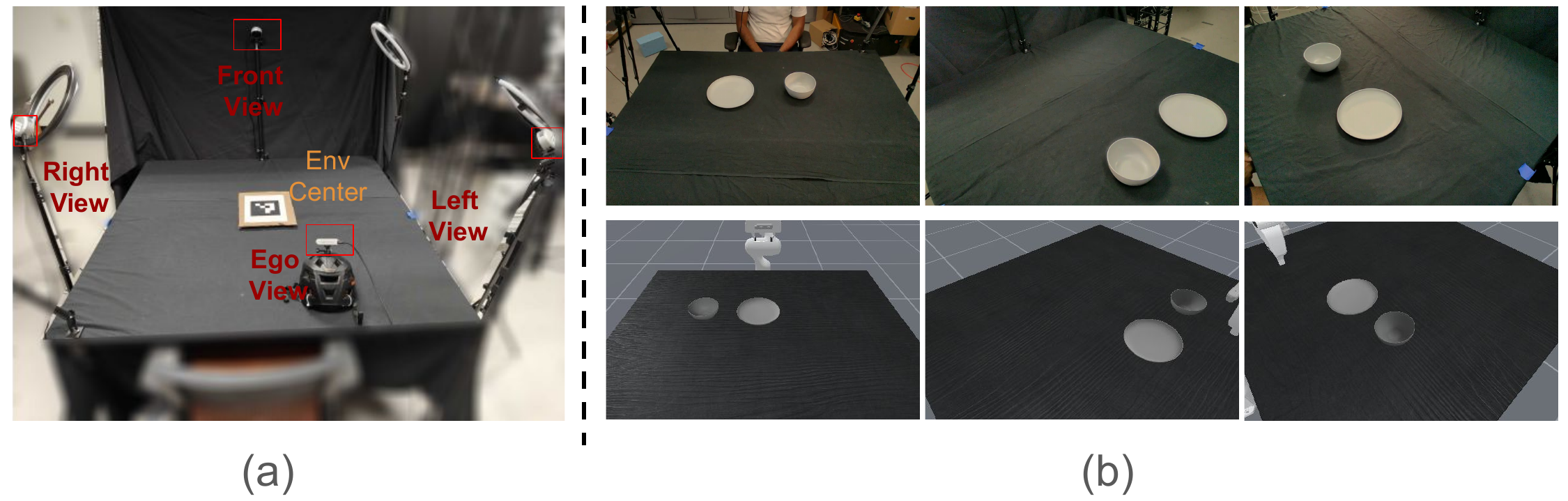}
    \caption{This figure shows the camera settings of the PD test suit($\diamondsuit$).(a) The four cameras in the real-world data collection setting. The ArUco marker marks the center of the table as the origin. (b) The corresponding camera views in the real world and simulation. From left to right we show the view from the front camera, the left camera, and the right camera. The top row shows the camera views from the real world, and the bottom row shows the camera views in the simulation. We compute the 3D camera poses with respect to the origin in the real world, and place the cameras at the corresponding poses in the simulation.}
    \label{fig:objects}
\end{figure}

Table~\ref{tab:task_descriptions} provides a natural language description for each task, along with the details of the implementation of the success criteria in the simulated environment.
Fig.~\ref{fig:start_and_end_conditions} shows the start and end images in the non-calibrated simulation environment. The top half of the figure shows the images of the initial states and the bottom half shows the last states of a successful execution. 
Fig.~\ref{fig:corresponding_frames} shows frames from the original human videos, human videos with keypoint annotations, the non-calibrated task environments, and the calibrated task environments from top to bottom, respectively.
Fig.~\ref{fig:objects} shows the images of the objects we use to collect real-world human videos. The first part of the figure shows the image of the task-relevant objects and the second part shows the distraction objects.

% \paragraph{Creating Simulation Environments with Calibrated Action Space.}
% \wgnote{Anant and Anmol: Describe the calibration steps in text. Table size measurement, finding the 3d pose of each cameras etc...}

\paragraph{Keypoint Annotation for Human Videos.}
Human demonstrations are processed through DIFT~\cite{tang2023emergent} to perform semantic object localization on the first frame of each demonstration, while CoTracker~\cite{karaev2024cotracker} propagates these keypoints to produce continuous keypoint trajectories throughout the demonstration.
We apply a low-pass filter to the recovered hand poses to reduce articulation-induced noise and suppress 180° orientation flips that the rigid transformation can introduce on non-rigid hand landmarks.
In the pose-defined setting where the action space of the simulation environment is calibrated to the real-world environment, these extracted 3D keypoint trajectories can be directly mapped to the end-effector keypoints of the robot in the simulation environments. We use these trajectories as the action annotations to train our pose-defined baseline. 

\subsection{Creating Paired Task in Simulation.}
\label{appendix:paired_sim}
To construct a faithful digital twin of every real-world task, we ported each tabletop setup into ManiSkill~3 / SAPIEN with a 7-DoF Franka Panda arm. Three design objectives guided the construction: (i) \emph{visual fidelity}, so that policies conditioned on real human videos see matching colors, shapes, and scene layouts at evaluation time; (ii) \emph{geometric and kinematic fidelity}, so that contact-rich behaviors transfer; and (iii) \emph{procedural reproducibility}, so that anyone can regenerate the paired robot data with a single command. The following paragraphs describe each component of this pipeline.

\paragraph{Asset reconstruction and import.} Rigid-body assets that appear in the real-world rig (cups, bowls, plates, lids, toys, the toaster cube, the cube, and the bin) are reconstructed as digital twins with SAM3D-Objects directly from RGB captures of the physical objects. Each reconstructed asset is exported as a triangle mesh for rendering and a V-HACD convex decomposition for collision, allowing accurate physics while preserving fine visual detail (color, texture, decals). Articulated assets, namely the drawer used in \texttt{CloseDrawer} and \texttt{ObjectInDrawer} and the toaster used in \texttt{Toaster}, are pulled directly from PartNet-Mobility, selecting in each case the asset that most closely resembles the corresponding real-world object in appearance and joint structure, so that the joint axes, joint limits, and link hierarchies are physically grounded. For every asset we precompute an upright resting orientation, an axis-aligned half-size, and a footprint radius (stored as \texttt{orientation\_result.json}), which the spawner uses to seed each episode with stable, collision-free initial poses.
Our digital twins target visual and geometric fidelity rather than exact dynamics. 
Rigid objects match real-world dimensions, with standardized mass ($300$ g) and friction ($0.5$).
For the drawer and toaster, we use the closest matching model from asset library, as multiple generation approaches failed to produce sufficiently reliable articulation and geometry.

\paragraph{Scene and camera calibration.} The simulated table inherits the dimensions of the physical table. Cameras are placed using the same ArUco-based extrinsics solved on the physical rig: a \texttt{front}/\texttt{front} camera at $(0.4, 0, 0.6)$, side cameras at $(0.4, \pm 0.4, 0.6)$, and an \texttt{overhead} camera at $(0, 0, 0.6)$, all looking at the table center. Each sensor renders $1280{\times}720$ RGB at $30$~fps with the respective FoV, mirroring the real-world stream that the LfO models consume.

\paragraph{Procedural trajectory generation.} Robot trajectories are produced by task-specific solvers built on top of \texttt{mplib}'s RRT-Connect and screw-motion planners (wrapped in \texttt{FixedPandaMotionPlanningSolver}). For \emph{pick-and-place} tasks (\texttt{PickCube}, \texttt{PutBasket}, \texttt{BowlOnPlate}, \texttt{StackCups}, \texttt{ObjectInDrawer}), GraspGen runs on per-object point clouds extracted from the current scene to propose candidate grasps; candidates are then re-ranked by a weighted cost combining angular distance from the current TCP, deviation from a downward approach, and joint-space displacement. The chosen grasp is executed through a five-phase template, pre-grasp, grasp, transport, pre-place, drop, with randomized standoffs along the approach axis to inject grasp diversity. For \emph{articulated} tasks (\texttt{CloseDrawer}, \texttt{Toaster}), the solver reads the live joint axis from the PartNet-Mobility model and waypoints the end-effector along that axis with a fixed standoff. Each trajectory is verified against the env's \texttt{evaluate()} success predicate before recording.

\paragraph{Distractor and texture randomization.} 
\texttt{DistractorMixin} class is composed into every environment and, when \texttt{distract=True} is called, it integrates 1-2 kinematic distractor assets per episode from an asset library of 5 task-irrelevant assets (banana, bowl, mouse box, orange, soda can), placing them by rejection sampling with a 25~cm clearance from any task-relevant object. To support deterministic replay of evaluation banks, the spawner also honors an explicit \texttt{distractor\_names} override so that recorded \texttt{state\_dict}s reproduce the exact same asset identities.

\paragraph{Data format and reproducibility.} Each successful trajectory is logged via \texttt{RecordEpisode} as a single HDF5 holding 36-dimensional proprioceptive observations, 8-dimensional absolute \texttt{pd\_joint\_pos} actions, and the full simulator state, together with five synchronized $512{\times}512$ MP4 streams (left, right, overhead, front and wrist). Table~\ref{tab:paired_sim_tasks} summarizes the solver strategy and primary asset source per task.

\begin{table}[h]
\centering
\small
\setlength{\tabcolsep}{4pt}
\begin{tabular}{lllc}
\toprule
\textbf{Task} & \textbf{Solver strategy} & \textbf{Primary asset source} & \textbf{Articulated} \\
\midrule
PushCube           & Scripted contact + screw motion             & Procedural primitive            & \xmark \\
PickCube           & GraspGen + 5-phase plan                     & Procedural primitive            & \xmark \\
BowlOnPlate        & GraspGen + 5-phase plan                     & SAM3D-Objects                   & \xmark \\
StackCups          & GraspGen + sequential 5-phase plan          & SAM3D-Objects                   & \xmark \\
PutBasket          & GraspGen + sequential drop into bin         & SAM3D-Objects                   & \xmark \\
EmptyBasket        & GraspGen + sequential lift-and-place        & SAM3D-Objects                   & \xmark \\
ObjectInDrawer     & GraspGen + screw motion along joint axis    & SAM3D-Objects + PartNet-Mobility & \cmark \\
CloseDrawer        & Waypoints along live prismatic joint axis   & PartNet-Mobility                & \cmark \\
Toaster            & Waypoints along live prismatic joint axis   & PartNet-Mobility                & \cmark \\
OpenLid            & Contact-based screw motion                  & SAM3D-Objects                   & \xmark \\
\bottomrule
\end{tabular}
\caption{Per-task summary of the paired simulation pipeline. Every task is implemented with the same Franka Panda arm, table geometry, and camera calibration as the real-world rig; only the solver template and asset source vary.}
\label{tab:paired_sim_tasks}
\end{table}

\subsection{Evaluation Protocol}
\label{appendix:eval_protocol}
We train one model checkpoint for each model class under each configuration.
Our experiment results are reported from three evaluation runs, and each run uses a different set of random seeds to create the evaluation environments.
The success rate for each individual evaluation run is computed from $20$ rollouts with different seeds. We report the means and standard error of means~(SEM) of the three evaluation runs for each model in each table.
More specifically, the number of each grid of each table is computed from $$3 \texttt{ evaluation runs } \times 20 \texttt{ rollouts per evaluation runs} = 60 \texttt{ total rollouts.}$$

\paragraph{Pre-recorded evaluation banks.}
To guarantee reproducibility, we collect $100$ environment seeds for each task under each configuration, and we use the first $60$ seeds to spawn the evaluation environments.
Across the evaluation runs,  we use the same $20$ human demonstrations unseen at training but distinct environment seeds.
For every \texttt{(task, env\_distract)} pair we pre-record an
\emph{evaluation bank} of $300$ entries and ship it with the benchmark.
Each entry stores (i) the ManiSkill RNG seed used to initialise the
episode, (ii) a full \texttt{state\_dict} snapshot capturing all object
poses, robot joint positions, and velocities at that initial configuration,
and (iii) for the with-distraction banks, the identities of the sampled
distractor objects so they can be reproduced exactly.
Entries are collected by an \emph{oracle-filtered} procedure: the
environment is reset with a fresh seed, the physics are allowed to settle,
and the state is snapshotted; a task-specific oracle solver then attempts
the task from that state. Only episodes in which the oracle \emph{succeeds}
are admitted to the bank --- failed configurations are discarded and a new
seed is tried. This guarantees that every bank entry is a genuinely solvable
starting configuration.
At evaluation, the protocol driver \emph{replays} bank entries rather than
re-seeding the env, which means:
\begin{enumerate}[leftmargin=5mm,nolistsep]
    \item Two different models see numerically identical starting states
    on every rollout, removing any variance from RNG-driven object spawns
    or distractor sampling.
    \item For the with-distraction cells, the bank also pins the distractor
    \emph{identities} via the \texttt{distractor\_names} override
    (Sec.~\ref{appendix:paired_sim}), so the same set of clutter objects
    appears regardless of which model is being evaluated.
    \item Banks are larger than any single pass needs, which allows the
    three-pass procedure described below to slice them into disjoint
    sub-banks.
\end{enumerate}

\paragraph{Evaluation Details for ND($\spadesuit$), WD($\heartsuit$), and ED($\clubsuit$).}
The models for these three test suites use human videos as inputs during evaluation. For each task, all models are trained using $80$ of the $100$ human demonstration videos and $100$ robot trajectories, with the remaining $20$ human demonstration videos held out for evaluation.
During training, human demonstrations and robot trajectories are paired within their corresponding task-specific pools rather than one-to-one, and therefore do not necessarily share the same object placements. During evaluation, the same $20$ held-out human demonstrations are used across the three evaluation runs and are paired with different initial environment configurations sampled from the corresponding task-specific evaluation pool.

For \textbf{ND($\spadesuit$)}, both the human videos and robot trajectories used for training contain no distractions; during evaluation, both the human videos and rollout environments contain no distractions. For \textbf{WD($\heartsuit$)}, both the human videos and robot trajectories used for training contain distractions; during evaluation, both the human videos and rollout environments contain distractions. For \textbf{ED($\clubsuit$)}, both the human videos and robot trajectories used for training contain no distractions; during evaluation, both the human videos and rollout environments contain distractions.

\paragraph{Evaluation Details for PD($\diamondsuit$).}
The model from the PD test suit does not take human videos as inputs for inference. We use all the $100$ human videos with keypoint annotations for training. During evaluation, the model only receives inputs from the evaluation environments.

\section{Extended Related Work}
\label{appdx:detailed_related_work}

\textbf{Human Action Video Datasets.}
The computer vision community has long been interested in understanding human action videos that can even have paired language commands with semantic annotations for actions taken by humans~\citep{ucf101, somethingsomething, activitynet,howto100M,hmdb}. 
These datasets have paired language commands with semantic annotations for actions taken by humans as well~\citep{hmdb, ucf101, kinetics,activitynet,somethingsomething}. 
Prior work~\citep{hmdb, ucf101, kinetics} frames action understanding from human activity videos as a classification problem, assigning a single action label to each video clip.
ActivityNet~\cite{activitynet} introduce semantic taxonomy and include temporally localized activity labels for untrimmed videos, allowing models to learn action labels of different levels of granularity, and the temporal relationship between action segments. 
As the vision models scale up, other annotations such as object movements~\cite{somethingsomething}, personal bounding boxes and atomic action labels~\cite{ava}, and language instructions~\cite{howto100M} are introduced to enhance models' ability to understand the spatio-temporal relationship of actions.
Egocentric datasets such as Assembly101~\cite{assembly101}, Epickitchen\cite{epickitchens} or Ego-Exo4D~\cite{egoexo4d} capture long-horizon, unscripted daily activities from the first-person viewpoints. Unfortunately, none of this data allows accurate comparisons for robotics community due to the lack of a paired environment for the robot to learn actions or skills.
Ego-Exo4D~\cite{egoexo4d} provides synchronized first-person and third-person videos for human activities, along with fine-grained action annotations, including language, pose, gaze, audio, keysteps, and expertise level. This dataset provides the foundation to train and evaluate foundation models on human action understanding.
These human action dataset from the CV community are designed to train and evaluate vision and language models on reasoning over human activities, but they do not care for whether the actions in the videos can be learned and reproduced by an embodied agent in a different environment.

\textbf{Learning from Observation Methods.}
Based on different knowledge transfer pathways, existing human-video-based techniques can roughly be organized into three different categories~\cite{lfosurvey}, including task-oriented transfer~\citep{seedo, physbrain, bcz, llmactionplanning}, observation-oriented transfer~\citep{vid2robot, r3m, mvp, mimicvideo}, and action-oriented transfer~\citep{mimicdroid, pokenet, motionTracks, rpx, pointpolicy, uniskill, rhyme}.
% Task-oriented Transfer
\textit{Task-oriented transfer} extracts semantic task information from human demonstration videos, then uses the extracted task information to control robot policies to perform the same tasks. 
There are two major lines of work for task-oriented transfer, including explicit task structure transfer, and implicit task intent transfer.
Explicit task structure transfer methods decompose a human demonstration video into temporal language instructions, and implicit task intent transfer methods learn to extract latent task intent representation by aligning human demonstration video with robot trajectories or natural language instructions by task intents~\citep{ bcz, xskill}.
% Observation-oriented Transfer
\textit{Observation-oriented transfer} focuses on directly bridging the visual gap between the human demonstration videos and the perception of the robot embodied agents. 
This line of work either directly transforms human demonstration videos into embodiment-agnostic or robot-like visual formats in the pixel space~\citep{h2r, mimicdreamer}, or learn embodiment-invariant visual embeddings to align the observation space between human and robot~\cite{mvp, vid2robot, r3m}.
% Action-oriented Transfer
\textit{Action-oriented transfer} aims to distill the action information, such as hand affordances and object affordances, and uses the action information to either train or guide robot policies. 
Similar to the other transfer mechanism, action-oriented transfer can be divideed into explicit affordances transfer and implicit latent actions transfer.
Explicit affordances transfer extracts affordances, such as 3D hand-positions~\citep{rpx, motionTracks, mimicdroid} and object geometric~\cite{pokenet}, from the human demonstration videos, and uses these actions to directly control the robot~\citep{rpx, pokenet, pointpolicy} or to train robot policies~\citep{motionTracks, mimicdroid}. 
This line of work requires calibration of the observation spaces and the action spaces between the human demonstrator and the embodied agents to produce any meaningful downstream robot policy.
On the other hand, implicit action transfer learns compact and embodiment-agnostic action priors directly from human videos by training forward dynamic models (FDMs) and inverse dynamic models (IDMs) using future observation reconstruction~\citep{uniskill, lapa, univla, rhyme}.
% Close it up!
As a result of the significant variance in assumptions, architectures, and evaluations of these methods, there is not a unified benchmark for systematic evaluation on the cross-embodiment learning from observation problem.
We aim to address this gap by constructing this benchmark, and conducting experiments with selected model from each of these transfer mechanisms.

\end{document}